\documentclass{article}

\usepackage{arxiv}

\usepackage[utf8]{inputenc} % allow utf-8 input
\usepackage[T1]{fontenc}    % use 8-bit T1 fonts
\usepackage{hyperref}       % hyperlinks
\usepackage{url}            % simple URL typesetting
\usepackage{booktabs}       % professional-quality tables
\usepackage{amsfonts}       % blackboard math symbols
\usepackage{nicefrac}       % compact symbols for 1/2, etc.
\usepackage{microtype}      % microtypography
\usepackage{lipsum}
\usepackage{graphicx}
\graphicspath{ {./images/} }
\usepackage{mathtools}

\newtheorem{definition}{Definition}

\newtheorem{proposition}{Proposition}

\usepackage{graphicx}
\usepackage{caption}
\usepackage{subcaption}
\usepackage{enumerate}% http://ctan.org/pkg/enumerate
\usepackage{multirow}
\usepackage{color,soul}     % hyperlinks
\hypersetup{
	colorlinks = true,
    urlcolor  = cyan,
	citecolor = blue,
	linkcolor = blue}

\title{Gaussian Experts Selection using Graphical Models}
\author{%
  Hamed Jalali  \\
  Department of Computer Science\\
  University of Tuebingen\\
  \texttt{hamed.jalali@wsii.uni-tuebingen.de} \\
   \And
   Martin Pawelczyk \\
   Department of Computer Science\\
   University of Tuebingen\\
  \texttt{martin.pawelczyk@uni-tuebingen.de} \\

   \AND
  Gjergji Kasneci \\
  Department of Computer Science\\
  University of Tuebingen\\
  \texttt{gjergji.kasneci@uni-tuebingen.de} \\

}

\begin{document}
\maketitle
\begin{abstract}
Local approximations are popular methods to scale Gaussian processes (GPs) to big data. Local approximations reduce time complexity by dividing the original dataset into subsets and training a local expert on each subset. Aggregating the experts' prediction is done assuming either conditional dependence or independence between the experts. Imposing the \emph{conditional independence assumption} (CI) between the experts renders the aggregation of different expert predictions time efficient at the cost of poor uncertainty quantification. On the other hand, modeling dependent experts can provide precise predictions and uncertainty quantification at the expense of impractically high computational costs. By eliminating weak experts via a theory-guided expert selection step, we substantially reduce the computational cost of aggregating dependent experts while ensuring calibrated uncertainty quantification. We leverage techniques from the literature on undirected graphical models, using sparse precision matrices that encode conditional dependencies between experts to select the most important experts. Moreover, our approach also provides a solution to the poor uncertainty quantification in CI-based models. 
\end{abstract}

\section{Introduction}
Gaussian processes \cite{Rasmussen} are interpretable and powerful statistical methods to deal with uncertainty in prediction problems. These non-parametric methods apply Bayes' theorem to discover complex linear and non-linear structures without the need for restrictive assumptions on the model. Due to their capabilities, they are widely used in practical cases, e.g. optimization \cite{Shahriari}, data visualization, and manifold learning \cite{Lawrence}, reinforcement learning \cite{Deisenroth2013}, multitask learning \cite{Alvarez}, online streaming models \cite{Huber, Le} and time series analysis \cite{Petelin, Tobar}.

The main limitation of these models is their computational cost on data sets with large numbers of observations. For a data set of size \textit{n}, the time and space complexities during training are $\mathcal{O}(n^3)$ and $\mathcal{O}(n^2)$, respectively. These high complexities are due to the inversion of the $n \times n$ kernel matrix and determinant computation. On the test set, the predictions require additional time and space complexities of $\mathcal{O}(n \log n)$. This issue restricts GPs to relatively small training data sets of the order of $\mathcal{O}(10^4)$. 

A popular method to scale GPs is the common divide-and-conquer approach. The idea is to partition the training data, perform local inference for each part separately, and combine the results to obtain a global posterior approximation  \cite{Tresp2001,Hinton,Cao,Deisenroth}. This distributed training approach generally imposes a \emph{conditional independence assumption} (CIA) between partitions, which allows factorizing the global posterior distribution as a product of local distributions. Although this assumption reduces the computational cost of GPs, it leads to statistical inconsistency \cite{Bachoc, Szabo}. That is, in the presence of infinite data, CIA based posterior approximations do not converge to the full GP posterior. Moreover, these approaches only provide pointwise variance and confidence intervals instead of a full GP predictive distribution. Solutions, that cope with the statistical consistency problem, suffer from extra computational costs \cite{Rulliere,Liu2018}. Hence, these solutions are impractical for large data sets.

Unlike the divide-and-conquer approach, which uses the full training set, another strategy tries to use a small part of the training data. For instance, the sparse approximation approach employs a subset of the data (called inducing points) and \textit{Nyström} approximations to estimate the posterior distribution \cite{Titsias, Hensman,Cheng,Burt}. Although this line of work provides a full probabilistic model, its capacity is restricted by the number of inducing points \cite{Bui, Moore}, and therefore it is not appropriate for large and high dimensional data sets.

In this paper, we propose a new expert selection approach for distributed GPs which improves the prediction quality in both dependent and independent experts aggregation methods. The full experts set is divided into two subsets, \emph{important} and \emph{unimportant} experts. The importance of an expert is measured according to his interactions with other experts. Particularly, we use a Markov random field to construct a sparse undirected graph of experts that provides information about the overall interactions. Experts with more significant connections are treated as important experts and are used for the final aggregation. Furthermore, relative to consistent aggregation methods that use dependency information, our approach also yields consistent predictions and substantially speeds up the running time, while maintaining competitive prediction performance. 

The structure of the paper is as follows. Section 2 introduces the problem formulation and related works. In Section 3 the proposed model and the inference process are presented. Section 4 shows the experimental results and we conclude in Section 5.

\section{Background and Problem Set-up}

\subsection{Gaussian Process}
Let us consider the regression problem $y=f(x)+\epsilon$, where $x\in \mathbb{R}^D$ and $\epsilon \sim \mathcal{N}(0,\sigma^2)$ which corresponds to the Gaussian likelihood $p(y|f)=\mathcal{N}(f, \sigma^2 I)$. The objective is to learn the latent function \textit{f} from a training set $\mathcal{D}=\{X,y\}$ of size $n$. The Gaussian process regression is a collection of random variables of which any finite subset has a joint Gaussian distribution. The GP then describes a prior distribution over the latent functions as $f \sim GP\left(m(x),k(x,x') \right)$, where $m(x)$ is a mean function and $k(x,x')$ is the covariance function (kernel) with hyperparameters $\psi$. The prior mean is often assumed to be zero, and the kernel is the  well-known squared exponential (SE) covariance function equipped with automatic relevance determination (ARD), 
\[k(x,x')=\sigma_f^2 \; \exp\left( -\frac{1}{2} \sum_{d=1}^D \frac{(x_d-x_d')^2}{\mathcal{L}_d} \right),\]
where $\sigma_f^2$ is the signal variance, and $\mathcal{L}_d$ is an input length-scale along the $d$-th dimension, and $\psi=\{\sigma_f^2,\mathcal{L}_1,\ldots,\mathcal{L}_D\}$. To train the GP, the hyperparameters $\theta=\{\sigma^2, \psi\}$ should be determined such that they maximise the log-marginal likelihood,
\begin{equation} \label{eq:1}
\log p(y|X)=-\frac{1}{2}y^T\mathcal{C}^{-1}y - \frac{1}{2} \log|\mathcal{C}|- \frac{n}{2} \log2\pi,
\end{equation}
where $\mathcal{C}=K+\sigma^2I$ and $K=k(X,X)$. According to \eqref{eq:1}, the training step scales as $\mathcal{O}(n^3)$ because it is affected by the inversion and determinant of the $n \times n$ matrix $\mathcal{C}$. Therefore, for large data sets, GP training is a time-consuming task and imposes limitations on the scalability of GPs.

\subsection{Local Gaussian Process Experts}
The local approximation Gaussian process uses the fact that the computations of the standard GP can be distributed among individual computing units. To do that, one divides the full training data set $\mathcal{D}$ into $M$ partitions (called experts) and trains standard GPs on these partitions. Let  $\mathcal{D}^{'}= \{\mathcal{D}_1,\ldots,\mathcal{D}_M\}$ be the partitions, and $X_i$ and $y_i$ be the input and output of partition $\mathcal{D}_i$. For a test set $X^*$, the predictive distribution of the $i$-th expert $\mathcal{M}_i$ is $p_i(y^*|\mathcal{D}_i,X^*)\sim \mathcal{N}(\mu_i^*,\Sigma_i^*)$, where its mean and covariance are respectively:
\begin{align}
\mu_i^* &= k_{i*}^T(K_i+\sigma^2I)^{-1}y_i,\label{eq:2} \\
\Sigma_i^* &= k_{**} - k_{i*}^T(K_i+\sigma^2I)^{-1}k_{i*}, \label{eq:3}
\end{align}
where $K_i=k(X_i,X_i)$, $k_{i*}=k(X_i,X^*)$, and $k_{**}=k(X^*,X^*)$. Aggregation is based on the assumption that the experts are conditionally independent leading to predictive distribution of the form 
$ p(y^*|\mathcal{D},X^*) \propto \prod_{i=1}^M p_i(y^*|\mathcal{D}_i,X^*)$.

The most popular aggregation methods are product of experts (PoE) \cite{Hinton} and Bayesian committee machine (BCM) \cite{Tresp}. Generalised product of experts (GPoE) \cite{Cao} and robust Bayesian committee machine (RBCM) \cite{Deisenroth} are more recent modified versions of PoE and BCM, which add the experts' weight quantifying each expert's contribution to the predictive distribution. Generalized robust Bayesian committee machine (GRBCM) \cite{Liu2018} is the most recent model, which uses a random subset $\mathcal{D}_b$ drawn from the entire training data and treats it as a global communication expert to augment each partition $\mathcal{D}_i$.\footnote{A brief overview of independent Gaussian experts models can be found in Appendix \ref{sec:DGP}.}

\subsection{Dependencies Between Experts' Predictions} \label{sec:2.3}
Assume the Gaussian experts $\mathcal{M}=\{\mathcal{M}_1,\ldots,\mathcal{M}_M\}$ have been trained on different partitions and the first and second moments of their local posterior distributions have been defined in \eqref{eq:2} and \eqref{eq:3}. Let $\mu^*(x^*)=[\mu_1^*(x^*),\ldots,\mu_M^*(x^*)]^T$ be an $M \times 1$ vector that contains the centered predictions of $M$ experts for a given test point $x^*\in X^*$. Assuming that $y_i$ in \eqref{eq:2} has not yet been observed, the authors of \cite{Rulliere}  considered the prediction mean $\mu^*_i(x^*)$ as a \emph{random variable}. This allows us to consider correlations between the experts' predictions and latent variable $y^*$, \ $Cov(\mu^*_i, y^*)$, and also leverage internal correlations between experts, \ $Cov(\mu^*_i,\mu^*_j)$ where $i,j=1,\ldots,M$. According to \eqref{eq:2}, the local experts are linear estimators: $\mu_i^*=\Gamma_i y_i$, where $\Gamma_i =k^T_{i*}(K_i + \sigma^2 I)^{-1}$. Using this result we can find the analytical expression for both covariances:
\begin{align}
Cov(\mu_i^*, y^*)=cov(\Gamma_i y_i,y^*)=\Gamma_i cov(y_i,y^*) =\Gamma_i k(X_i,X^*),\label{eq:4} \\
Cov(\mu_i^*, \mu_j^*)=cov(\Gamma_i y_i,\Gamma_j y_j)=\Gamma_i cov(y_i,y_j) \Gamma_j^T = \Gamma_i k(X_i,X_j) \Gamma_j^T. \label{eq:5}
\end{align}
%The inverse of $Cov(\mu_i^*, \mu_j^*)$ encodes the conditional dependencies between experts' predictions. Since the experts are Gaussian random variables, $Cov(\mu_i^*, \mu_j^*)^{-1}=0$ entails conditional independence between $\mu_i^*$ and $\mu_j^*$ given all other experts.  

\subsection{Nested Point-wise Aggregation of Experts (NPAE)} \label{sec:2.4}
Assume the means (predictions) of the local predictive distributions are random variables. Then $k_A(x^*)=Cov\left(\mu^*(x^*), y^*(x^*)\right)$ and $K_A(x^*)=Cov\left(\mu^*(x^*),\mu^*(x^*)\right)$ are the point-wise covariances. For \emph{each} test point $x^*$, $k_A(x^*)$ is a $M\times1$ vector and $K_A(x^*)$ is a $M\times M$ matrix and according to \eqref{eq:4} and \eqref{eq:5} their elements are defined as $k_A(x^*)_i=\Gamma_i k(X_i,x^*)$ and $K_A(x^*)_{ij}= \Gamma_i k(X_i,X_j) \Gamma_j^T$, where $\Gamma_i =k(X_i,x^*)^T(K_i + \sigma^2 I)^{-1}$ and $i,j=1,\ldots,M$. The task is to aggregate variables $\mu_i^*(x^*), i=1,\ldots,M$ into a unique predictor $y_A^*(x^*)$ of $y^*(x^*)$. 

As a consequence from the the choice of the prior, the joint distribution of random variables $(y^*,\mu_1^*,\ldots,\mu_M^*)$ is a multivariate normal distribution because any vector of linear combinations of observation is itself a Gaussian vector. This fact is used to define the aggregated predictor, but it also implies that the experts' predictions $(\mu_1^*,\ldots,\mu_M^*)$ follow a multivariate Gaussian. Employing properties of conditional Gaussian distributions for the centered random vector $\left(y^*(x^*),\mu^*(x^*)\right)$ allows for the following aggregation:

\begin{definition}[\textbf{Aggregated predictor}) \cite{Rulliere}]
For the test point $x^*$ and sub-model predictions $\mu_1^*(x^*),\ldots,\mu_M^*(x^*)$, the aggregated predictor is defined as
\begin{equation} \label{eq:6}
y_A^*(x^*)=k_A(x^*)^T K_A(x^*)^{-1}\mu^*(x^*).
\end{equation}
\end{definition}
In \cite{Rulliere,Bachoc} the authors showed that this linear estimator is the \textit{best linear unbiased predictor} (BLUP) where the invertibility condition on $K_A(x^*)$ can be avoided by using matrices pseudo-inverses. This aggregated predictor has Gaussian distribution and its moments can easily be calculated using $k_A(x^*)$ and $K_A(x^*)$. This method is known as the nested pointwise aggregation of experts (NPAE) and provides consistent predictions.

\subsection{Consistency} \label{sec.consistency}
The conventional PoE, GPoE, BCM, and RBCM produce inconsistent predictions and generally do not converge to the predictive distribution of the standard GP when $n \to \infty$ \cite{Bachoc,samo2016}. Using normalized equal weights, GPoE asymptotically converges to the full GP distribution, however, it is too conservative \cite{Szabo}. Integrating a global partition into RBCM, \cite{Liu2018} showed that GRBCM can provide consistent results. However, it still uses the conditional independence assumption between non-global experts, which sometimes yields poor results. 

To deal with the inconsistency issue, NPAE considers dependencies between individual experts' predictions and uses the property of conditional Gaussian distributions to find the predictive distribution of $y^*$. Although it theoretically provides consistent predictions, its aggregation step suffers from high time complexity because it needs to determine the inverse of a $M \times M$ covariance matrix between experts at \emph{each} test point $x^*$, i.e. $\mathcal{O}(n_t M^3)$, where $n_t$ is the number of available test points. This leads to impractically long running times for many partitions and large test sets.

To mitigate the drawbacks of the NPAE and CI-based aggregations, we will implicitly incorporate an expert selection step by using expert interaction strengths. The influences of the weak experts' prediction on prediction quality and computational cost can be eliminated by excluding them in both methods. In the next section, we present our new approach. First, we explain the proposed pruning step which is based on undirected graphical models. Second, we propose a new aggregation method with this pruning approach and consider its asymptotic properties.

\section{Expert Selection with Gaussian Graphical Models}

At the heart of our work is the following ingredient. We assume that the local experts can be divided into 2 sets: important and unimportant experts. Aggregation based on important experts provides consistent and competitive predictions with significantly reduced computational time. Instead of expert selection, state-of-the-art (SOTA) methods use expert weighting, i.e. they assign weights to each expert, but they do not provide a systematic way to determine the most important experts. In NPAE, the point-wise weights of local expert predictions are defined as $k_A(x^*)^T K_A(x^*)^{-1}$. In Appendix \ref{experts_weights}, we give a brief overview of how expert weighting is conducted for CI-based methods, and discuss the difference to our expert selection approach. In section \ref{sec:3.1}, we propose a novel approach that estimates the set of important experts based on the experts' interactions in an undirected graph.

\subsection{Gaussian Graphical Models for Correlated Gaussian Experts} \label{sec:3.1}
Since the experts' mean predictors are random variables with joint Gaussian distribution, we can study them using the machinery of sparse Gaussian graphical models (GGM). In GGMs, the assumption is that there exist few interactions, i.e., edges, between variables; a penalty parameter controls the network's sparsity. It will allow us to obtain a set of important experts $\tilde{\mathcal{M}}=\{\tilde{\mathcal{M}}_1,\ldots,\tilde{\mathcal{M}}_p\}$, where $p < M$.

\begin{definition}[\textbf{Important and Unimportant Experts}] \label{def.2}
Let $\mathcal{M}=\{\mathcal{M}_1,\ldots,\mathcal{M}_M\}$ be the set of GP experts and let $\mathcal{G}(\mathcal{V},\mathcal{E})$ be their Gaussian graph, where $\mathcal{V}=\mathcal{M}$ and the edges $\mathcal{E}$ are the interactions between experts. Then, $\tilde{\mathcal{M}}_{\alpha}$ is the set of \emph{important} experts, if it contains $\alpha \times 100$ of the most connected experts in $\mathcal{G}$, where $\alpha \in [0,1]$. Its complement, $\tilde{\mathcal{M}}^c_{\alpha}$, contains the remaining \emph{unimportant} experts and $\mathcal{M}=\tilde{\mathcal{M}}_{\alpha} \cup \tilde{\mathcal{M}}^c_{\alpha}$. The selection rate $\alpha$ controls how many experts should be assigned to $\tilde{\mathcal{M}}_{\alpha}$, i.e. $\vert \tilde{\mathcal{M}}_{\alpha} \vert = M_{\alpha}=\lceil\alpha M \rceil$, where $\vert \cdot \vert$ denotes the cardinality.   \end{definition}

\paragraph{Gaussian Graphical Models.}
Undirected graphical models (known as pairwise Markov random fields (MRF)) provide a framework for encoding joint distributions over large numbers of interacting random variables. This framework uses a matrix of parameters to encode the graph structure. In other words, it encodes the edges as parameters: if there is a connection between two nodes, then there is a non-zero parameter indicating that the edge between nodes is present in the graph. 
In Gaussian graphical models \cite{Rue,Uhler,Drton} the nodes are continuous Gaussian random variables. The main assumption underlying GGMs is that all variables follow a multivariate Gaussian distribution,

\begin{equation} \label{GGM_variance}
\begin{aligned}
 p(\mu^*|\xi, \Psi)= \frac{1}{(2\pi)^{M/2}|\Psi|^{1/2}}\exp\left\{-\frac{1}{2}(\mu^*-\xi)^T\Psi^{-1}(\mu^*-\xi) \right\},
\end{aligned}
\end{equation}

where $\mu^*=\{\mu^*_1,\ldots,\mu^*_M\}$ are the variables (nodes), M is the number of variables, and $\xi$ and $\Psi$ are the mean vector and the covariance matrix, respectively. We can rewrite \eqref{GGM_variance} using the precision matrix $\Omega$,   

\begin{equation} \label{GGM_precision}
\begin{aligned}
 p(\mu^*|\eta, \Omega)= \frac{|\Omega|^{1/2}}{(2\pi)^{M/2}} \exp\left\{-\frac{1}{2}(\mu^*-\xi)^T\Omega(\mu^*-\xi) \right\} \propto \exp\left\{-\frac{1}{2}{\mu^*}^T\Omega \mu^* + \eta^T \mu^* \right\} ,
\end{aligned}
\end{equation}

where $\Omega=\Psi^{-1}$ and $\eta=\Omega \xi$. The matrix $\Omega$ is also known as the potential or information matrix. Without loss of generality, let $\xi= 0$, then the distribution of a GGM shows the potentials defined on each node $i$ as $\exp\{-\Omega_{ii}(\mu^*_i)^2\}$ and on each edge $(i,j)$ as $\exp\{-\Omega_{ij} \mu^*_i \mu^*_j\}$. 

In the correlation network, expressed through \eqref{GGM_variance}, $\Psi$ encodes independences: if $\Psi_{ij}=0$, then $\mu^*_i$ and $\mu^*_j$ are assumed to be \emph{independent}. On the other hand, in the  GGM in \eqref{GGM_precision} we have: if $\Omega_{ij}=0$, then $\mu^*_i$ are $\mu^*_j$ are \emph{conditionally independent} given all other variables, i.e. there is no edge connecting $\mu^*_i$ and $\mu^*_j$ in the graph.\footnote{We would like to emphasize that this does not hold in general, but here it does hold since we assumed that all variables are jointly Gaussian distributed.}Besides, $\Omega_{ij} \approx 0$ means there is a weak interaction between $\mu^*_i$ and $\mu^*_j$.

\subsection{Network Learning for the Aggregated Posterior} \label{sec:3.2}
We use the GGM as follows: the locally trained experts represent nodes and network learning involves computing the precision matrix $\Omega$ that provides the interactions (edges) between the experts. To keep inference in GGMs tractable, one imposes a \emph{sparsity assumption}: there exist few edges in the network, and thus $\Omega$ is a sparse matrix. This assumption is empirically meaningful in an experts' network because the strong interaction of one expert can be typically limited to only a few other experts. In fact, since the local predictions of this expert are close to the local predictions of its adjacent experts, it has stronger interactions (or similarities) with them, see \cite{Luxburg} for more information about similarity and dissimilarity.

The literature has suggested several inference algorithms to recover the edge set. Lasso regression \cite{Hastie2015,Meinshausen2006} can be used to perform neighborhood selection to recover the network. 
\cite{Meinshausen2006} and \cite{Wainwright} proved that, under some regularity conditions, Lasso asymptotically recovers the correct and relevant subset of edges. The authors of \cite{Friedman2008} proposed a more efficient inference algorithm, the graphical Lasso (GLasso), which adopts a maximum likelihood approach subject to an $l_1$ penalty on the coefficients of the precision matrix. Its inference is fast and has been improved in subsequent works \cite{Banerjee,Friedman2010,Friedman2011,Hallac}. The authors in \cite{jalali2020} used the GLasso to detect dependencies between Gaussian experts and define clusters of strongly dependent experts. 

The GLasso objective is formalized as follows. Let $S$ be the sample covariance of local predictions $\mu^*$. Then, the Gaussian log likelihood of the precision matrix $\Omega$ is equal to $\log |\Omega| - trace(S \Omega)$. The Graphical Lasso maximises this likelihood subject to an element-wise $l_1$-norm penalty on $\Omega$. More precisely, the objective function is,
\begin{equation*} %\label{eq:10}
\widehat{\Omega}_{\lambda}= \arg\max_{\Omega} \log |\Omega| - trace(S \Omega) - \lambda \left\Vert \Omega \right\Vert_1,
\end{equation*}
where the estimated expert network is then given by the non-zero elements of $\widehat{\Omega}_{\lambda}$. 

\begin{definition}[\textbf{$\lambda$-related Expert Importance}] \label{def.3}
The $\lambda$-related importance of expert $\mathcal{M}_i$ based on its interactions is defined as $\mathcal{I}_i=\sum_{j=1, j\neq i}^M |\widehat{\Omega}_{\lambda,ij}|$, resulting in the sorted importance set $\mathcal{I}=\left\{\mathcal{I}_{i_1},\mathcal{I}_{i_2}, \ldots, \mathcal{I}_{i_M} \right\}$. 
\end{definition}

\begin{figure} [hbt!] 
\centering
\subcaptionbox{$\alpha=50\%$}{\frame{\includegraphics[width=0.32\columnwidth]{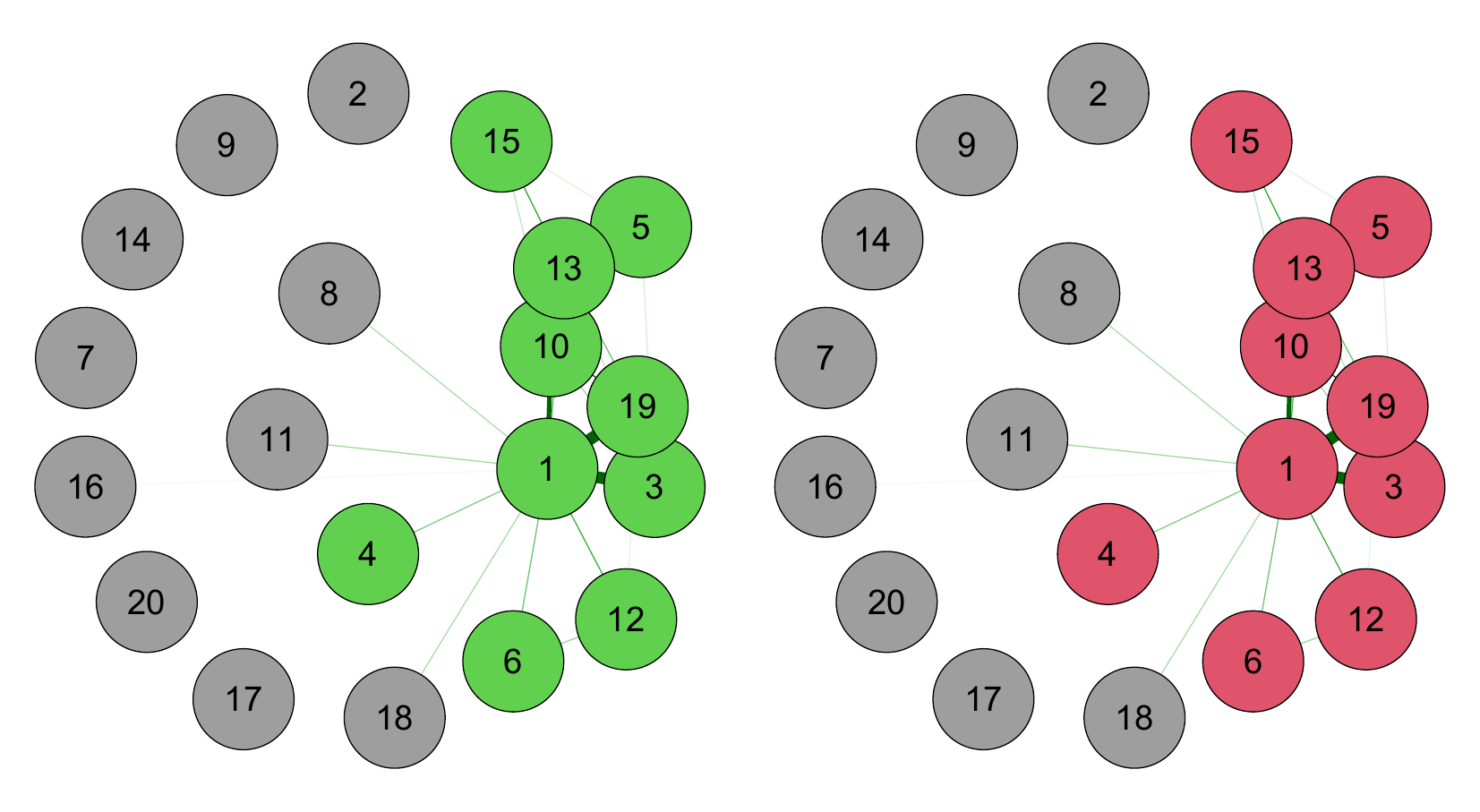}}}
\subcaptionbox{$\alpha=70\%$}{\frame{\includegraphics[width=0.32\columnwidth]{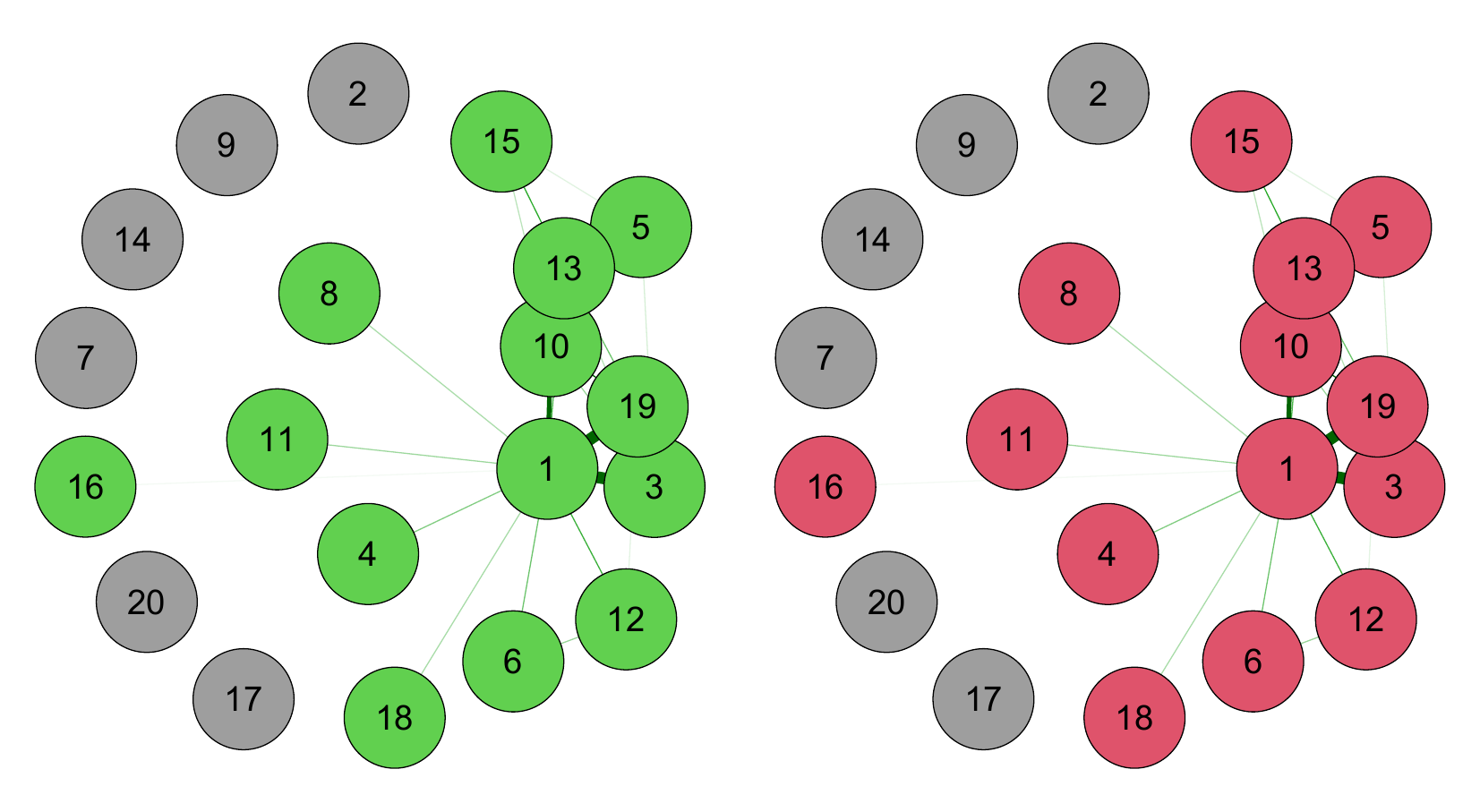}}}
\subcaptionbox{$\alpha=80\%$}{\frame{\includegraphics[width=0.32\columnwidth]{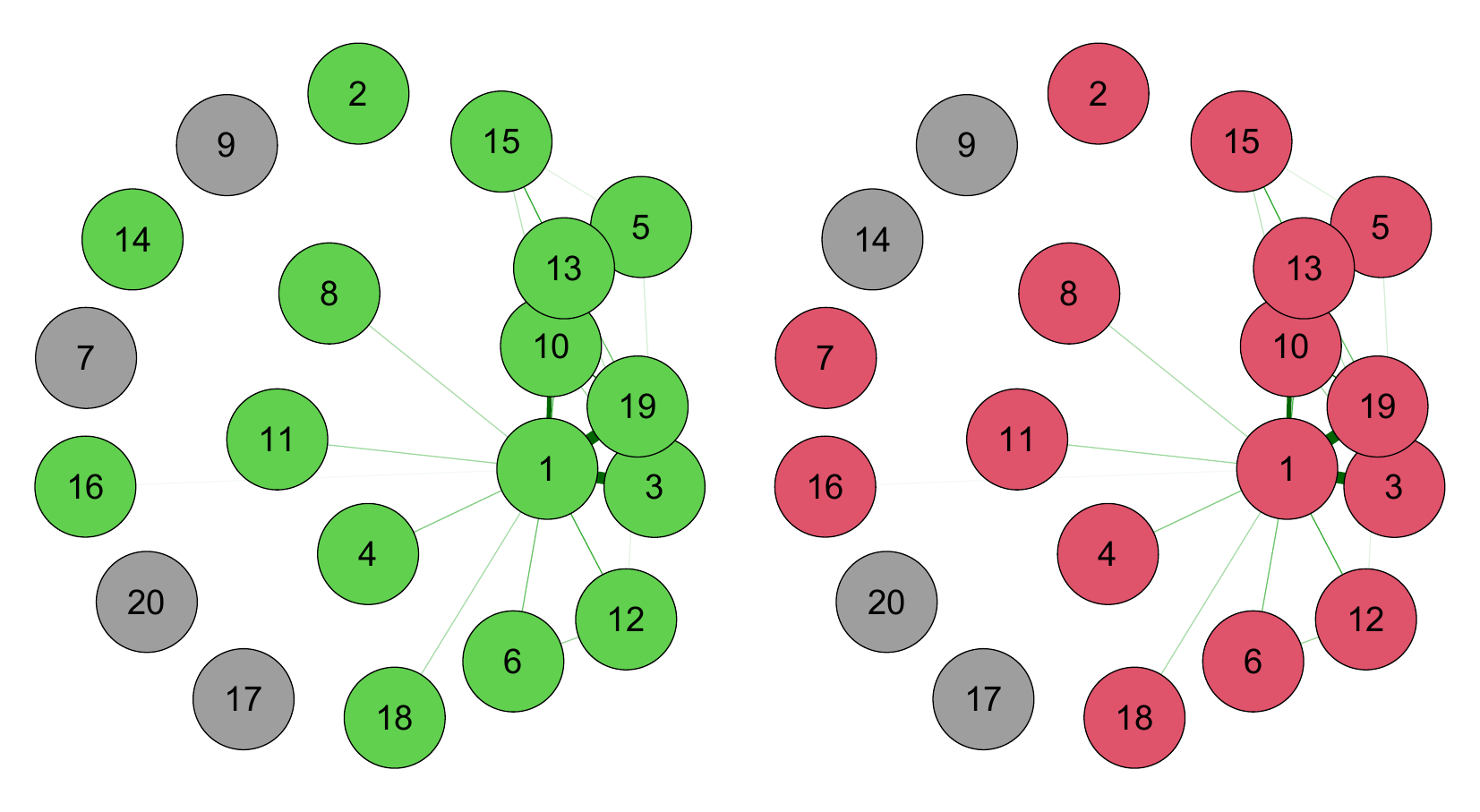}}}
%\subcaptionbox{$\alpha=75\%$}{\includegraphics[width=0.40\columnwidth]{net75.pdf}}
\caption{\textbf{Ablation experiment 1.} Expert selection for synthetic data from \eqref{f_x} with varying expert set strength $\alpha$ and $\lambda=0.1$. The green nodes reveal the $\alpha \%$ of the best best experts w.r.t. their individual MSE errors while the red nodes are the most important experts according to Definition \ref{def.3}.}
\label{fig.net}
\vskip -0.2in
\end{figure}

Figure \ref{fig.net} depicts graphs based on $5 \times 10^3$ training points and $500$ test points from \eqref{f_x} divided among 20 experts (250 training points for each expert). First, after training the local experts the prediction quality of each expert is measured based on their mean squared error (MSE). Then the experts that are among the $\alpha\%$ of the best experts with the lowest MSE values are highlighted in green. To evaluate the proposed expert selection approach, the most important experts -- based on their interactions with other experts -- are depicted as red nodes. For Figure \ref{fig.net} we have used the network inference method based on GLasso approximation \cite{Friedman2011} and Definition \ref{def.3}. It can provide a selection approach which tends to choose the best experts as the important experts.

\subsection{Point-wise Aggregation with Expert Selection} \label{sec:3.3}
Assume $\tilde{\mathcal{M}}_{\alpha}$ and $\tilde{\mathcal{M}}^c_{\alpha}$ are important and unimportant experts' sets as defined by Definitions \ref{def.2} and \ref{def.3}. Further, let $\tilde{\mu}_{\alpha}^*$ represent the local predictions of the experts in $\tilde{\mathcal{M}}_{\alpha}$. We use $\tilde{k}_{\alpha}$ and $\tilde{K}_{\alpha}$ to denote $Cov\left(\tilde{\mu}_{\alpha}^*, y^*\right)$ and  $Cov\left( \tilde{\mu}_{\alpha}^*,\tilde{\mu}_{\alpha}^*\right)$, respectively. 
Using Definition \ref{def.2}, the following proposition gives the predictive distribution of \textbf{NPAE*}, our new NPAE estimator with expert selection.

\begin{proposition}[\textbf{Predictive Distribution}]
Let $X$ be a compact, nonempty subset of $\mathcal{R}^{n \times D}$, $\mu^*(x^*)=[\mu_1^*(x^*),\ldots,\mu_M^*(x^*)]^T$ be the sub-models' predictions at a test point $x^*$. We further assume that (i) $\lim_{n\to \infty} M = \infty$ and (ii) $\lim_{n\to \infty} m_0 = \infty$, where $m_0$ is the partition size. The vector $\left(\mu_1^*,\ldots,\mu_M^*,y^*\right)$ is a multivariate Gaussian distribution and we obtain the following results:
\begin{enumerate}[(I)]
  \item The experts' importance-based aggregated estimator of $y^*(x^*)$ given $\mu^*(x^*)$ is Gaussian with mean and variance given by:
   \begin{align}
   \mathbb{E}[y^*(x^*)|\tilde{\mu}_{\alpha}^*(x^*)]  =y_{\alpha}^*(x^*)  
   = \tilde{k}_{\alpha}(x^*)^T \tilde{K}_{\alpha}(x^*)^{-1}\tilde{\mu}_{\alpha}^*(x^*)  \label{aggregate_mean} \\
    \mathbb{V}[y^*(x^*)|\mu^*(x^*)]  = k(x^*,x^*)   -\tilde{k}_{\alpha}(x^*)^T \tilde{K}_{\alpha}(x^*)^{-1} \tilde{k}_{\alpha}(x^*).
        \label{aggregate_variance}
    \end{align}

  \item $y_{\alpha}^*(x^*)$ is a consistent estimator for $y^*(x^*)$, i.e.  
      \begin{equation} \label{eq:consistent_prediction}
    \begin{aligned}
    \lim_{n\to \infty}\sup_{x^* \in X^*} \mathbb{E}\left[y^*(x^*)-y_{\alpha}^*(x^*)\right]^2\to 0.
    \end{aligned}
    \end{equation}
 \item $y_{\alpha}^*(x^*)$ provides an appropriate approximation for $y^*_A(x^*)$ in \eqref{eq:6}, i.e. 
   \begin{equation} \label{eq:npae_approximation}
    \begin{aligned}
    \lim_{n\to \infty} y_{\alpha}^*(x^*)\to  y_A^*(x^*).
    \end{aligned}
    \end{equation}
\end{enumerate}
\label{prop:predictive_distribution}
\end{proposition}
\paragraph{Proof.} See Appendix \ref{sec:proof}.

Proposition \ref{prop:predictive_distribution} reveals that the aggregation of the most important experts in NPAE* leads to consistent predictions and the estimator in \eqref{aggregate_mean} provides an appropriate approximation for the NPAE estimator in \eqref{eq:6}. This expert selection offers two interesting implications. First, it suggests that the unimportant experts are irrelevant to obtain consistent predictions and thus the prediction quality can suffer from these weak experts. Second, using only the most important experts indicates that computation time can be substantially reduced. In a nutshell, Proposition \ref{prop:predictive_distribution} suggests that the described expert selection strategy leads to a consistent and efficient estimator\footnote{In Appendix \ref{sec:Computational_Costs}, the computational cost of this expert selection method has been discussed}.

\subsection{CI-Based Models with Experts Selection}\label{sec:3.4}
Our selection strategy can easily be extended to CI-based methods. Although it can not improve the asymptotic properties of CI-based models, it improves upon the prediction quality of the baseline models. Thus, the modifications consist of excluding the weak experts based on the procedure explained in Section \ref{sec:3.1} which leads to (G)PoE* and (R)BCM* models.\footnote{Using the expert selection with GRBCM requires an extra step which we discuss in Section \ref{sec:GRBCM*}.}

\section{Experiments} \label{sec:4}
In this section, we evaluate the quality of the expert-importance-based estimator. We consider the prediction quality of our proposed model and other related SOTA models by using both simulated and real-world data sets. The quality of predictions is evaluated in two ways: we use the standardized mean squared error (SMSE) and the mean standardized log loss (MSLL). The SMSE measures the accuracy of the prediction mean, while the MSLL evaluates the quality of the predictive distribution \cite{Rasmussen}. In addition, the conventional mean absolute error (MAE) is also used in Section \ref{sec.4.1}. We use the standard squared exponential kernel with automatic relevance determination and a Gaussian likelihood. Since the disjoint partitioning of training data captures the local features more accurately and outperforms random partitioning \cite{Liu2018}, it is mostly used in our experiments.
%For all experiments the tuning parameter of the GLasso is set to $\lambda=0.1$.
All experiments have been conducted in MATLAB using the GPML package.\footnote{\url{http://www.gaussianprocess.org/gpml/code/matlab/doc/}} 

\subsection{Sensitivity Analysis using Synthetic Example} \label{sec.4.1}
The goal of our first experiment is to study the effect of expert selection on prediction quality and computation time. It is based on simulated data of a one-dimensional analytical function \cite{Liu2018},
\begin{equation}
f(x) = 5x^2sin(12x) + (x^3 -0.5)sin(3x-0.5)+4cos(2x) + \epsilon,  \label{f_x}
\end{equation}
where $\epsilon \sim \mathcal{N}\left(0, (0.2)^2\right)$. We generate $n=5\times 10^3$ training points in $[0,1]$, and $n_t=0.1n$ test points in $[-0.2,1.2]$. The data is normalized to zero mean and unit variance. We vary the number of experts,$M={10,20,30,40,50}$, to consider different partition sizes.   
$K$-means method is used for the partitioning to compare the prediction quality of NPAE* with its original version NPAE, GPoE, RBCM, GRBCM and the full GP. Since the quality of PoE and BCM methods are low, we ignore them in this part. For $\alpha$, we use two values, 0.5 and 0.8, which mean $50\%$ and $20\%$ of experts are excluded in the final aggregation.
 
\begin{figure}[hbt!] 
\centering
\subcaptionbox{\label{fig:expert_selection_mae}}{\includegraphics[width=0.24\columnwidth]{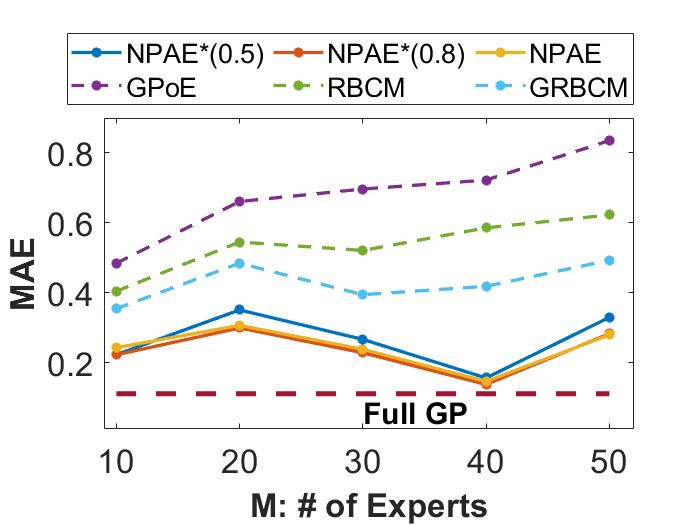}}\hfill
\subcaptionbox{\label{fig:expert_selection_smse}}{\includegraphics[width=0.24\columnwidth]{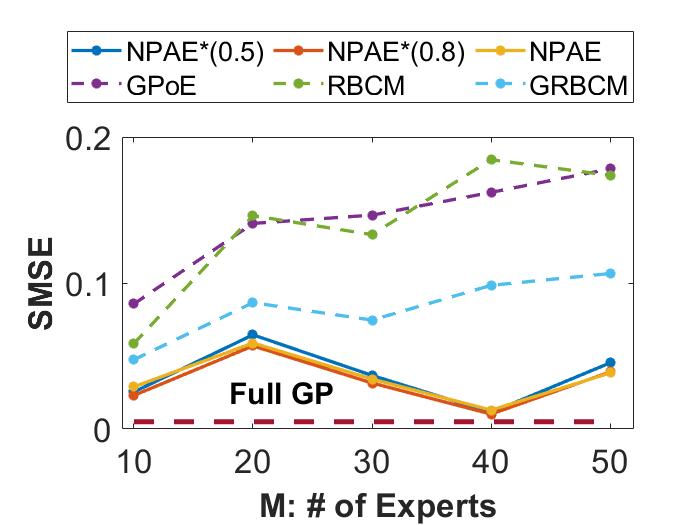}}\hfill
\subcaptionbox{ \label{fig:expert_selection_msll}}{\includegraphics[width=0.24\columnwidth]{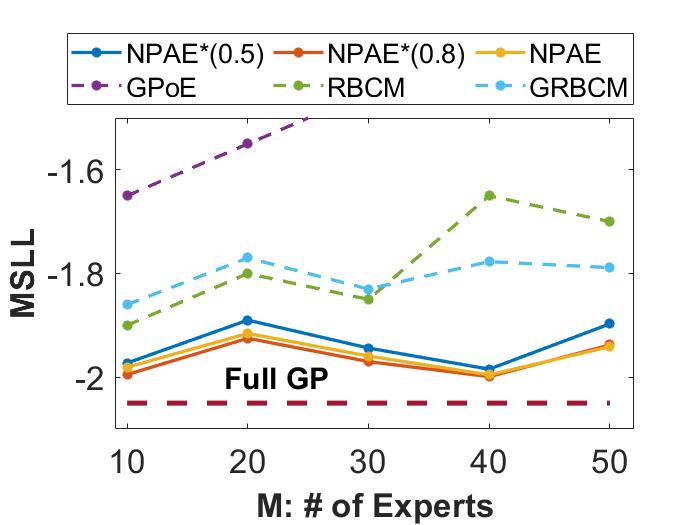}}\hfill
\subcaptionbox{ \label{fig:expert_selection_time}}{\includegraphics[width=0.24\columnwidth]{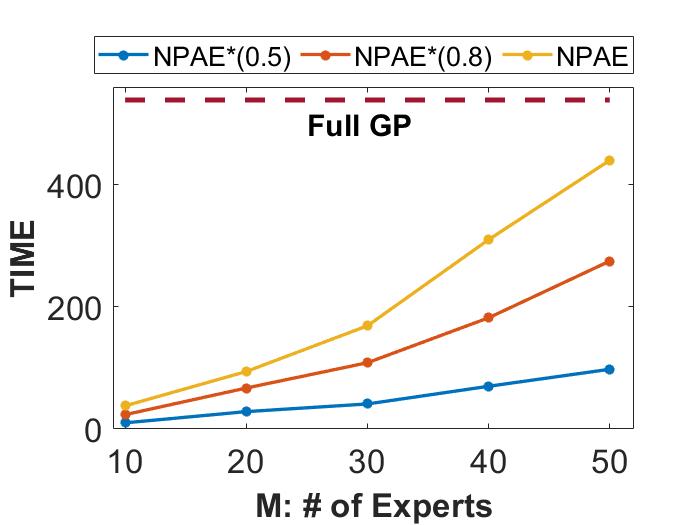}}\hfill
\caption{\textbf{Prediction quality and computation time} as a function of experts. MAE, SMSE, MSLL, and running time of NPAE* with $50\%$ and $80\%$ of experts ,NPAE, GPoE, RBCM, GRBCM, and full GP for $5 \times 10^3$ training points and partition size $M={10,20,30,40,50}$ with $K$-means partitioning. The x-axis shows the number of partitions (experts) ($M$) used in training step for fixed $\lambda=0.1$.}
\label{fig.prediction_quality_fx}
\vskip -0.1in
\end{figure} 

Figure \ref{fig.prediction_quality_fx} depicts the prediction quality of NPAE* compared to other baselines and the full GP. For NPAE*, we vary the percentage of selected experts which leads to NPAE*(0.5) and NPAE*(0.8). These two methods reflect the prediction quality of the approximation method, proposed in Proposition \ref{prop:predictive_distribution} for pointwise NPAE method. The plots in \ref{fig:expert_selection_mae}, \ref{fig:expert_selection_smse} and \ref{fig:expert_selection_msll} indicate that even 50 percent of the highly connected experts, as selected by NPAE*(0.5), provide accurate approximation for NPAE while NPAE*(0.8) even improves the prediction quality of NPAE to some extent. When the number of experts increases, the difference in prediction error between NPAE family and CI-based models increases, implying that with small partition size CI-based models are not capable to provide accurate predictions. Figure \ref{fig:expert_selection_time} depicts the trade-off between the inclusion of more experts and computation time in NPAE method. Remarkably, NPAE*(0.5) and NPAE*(0.8) provide competitive prediction quality compared to NPAE in just a fraction of NPAE's running time.

\subsection{Real-World Data Sets} \label{sec:4.2}

\subsubsection{Partitioning Strategies in a Medium-Scale Data Set} \label{sec:4.2.1}
In this section, we use a medium-scale real-world data set to assess the effect of the data assignment strategy on the prediction quality. The \textit{Pumadyn}\footnote{\url{https://www.cs.toronto.edu/~delve/data/pumadyn/desc.html}} is a generated data set with 32 dimensions and 7,168 training points and 1,024 test points. Both disjoint and random partitioning are used to divide the data set into 15 subsets. We consider the GPoE, RBCM, GRBCM, and NPAE with random and K-means partitioning. For the GPoE, RBCM, GRBCM and NPAE, we evaluate the proposed expert selection strategy with $\alpha= 0.8$ for GPoE, RBCM and GRBCM and $\alpha = 0.5$ and $\alpha = 0.8$ for NPAE. The penalty term $\lambda$ is 0.1 in this experiment.

Table \ref{table.pumadyn} depicts the prediction quality of local approximation methods for this data set. The column Type shows the interactions between experts in the aggregation method, D for dependent experts, and CI for conditionally independent experts. The GPoE*, RBCM*, and NPAE* are the modified versions of GPoE, RBCM, and NPAE, respectively. For CI-based methods, the modified methods outperform their original methods. For NPAE, the NPAE* with $50\%$ of the experts returns an appropriate approximation for NPAE which has been discussed in Proposition \ref{prop:predictive_distribution}. The NPAE* with $80\%$ of the experts excludes only weak experts which leads to a significant improvement in prediction quality and offers results very close to the original NPAE. Besides, it has been widely accepted that the local features of the data can be captured more accurately in disjoint partitioning, see \cite{Liu2018}. Table \ref{table.pumadyn} confirms this fact where the original and modified versions of aggregation methods reveal better prediction quality in K-means partitioning.

\begin{table}[hbt!]
\caption{SMSE and MSLL of different baselines on the \textit{Pumadyn} data set for different partitioning strategies. Both dependent (D) and conditionally independent (CI) aggregation methods are used.}
\label{table.pumadyn}
\vskip -0.3in
\begin{center}
\begin{small}
\begin{sc}
\begin{tabular}{lccccr}
\toprule
\multicolumn{1}{c}{} &
\multirow{2}{*}{} &
\multicolumn{2}{c}{K-means} &
\multicolumn{2}{c}{Random} \\
\toprule
 Model &Type & SMSE  & MSLL  & SMSE & MSLL   \\
\midrule
GPoE    &CI& 0.0483 & -1.5166& 0.0488& -1.5125  \\
\textcolor{blue}{GPoE*}  & CI & 0.0477 & -1.5213& 0.0485& -1.518  \\
RBCM    &CI & 0.0478 & 1.1224 & 0.0485& 2.9205 \\
\textcolor{blue}{RBCM*}    &CI & 0.0474 &0.3949& 0.0482& 1.7881  \\
GRBCM      &CI& 0.0499& -1.4949 & 0.0507 & -1.502  \\
\midrule
\textcolor{blue}{NPAE*(0.5)}    &D & 0.0470 &-1.5285& 0.0477 & -1.5265  \\
\textcolor{blue}{NPAE*(0.8)}  &D& \textbf{0.0468} & \textbf{-1.531} & \textbf{0.0474} &\textbf{-1.530}\\
NPAE    &D& \textbf{0.0466} & \textbf{-1.536} & \textbf{0.0472} &\textbf{-1.534}\\

\bottomrule
\end{tabular}
\end{sc}
\end{small}
\end{center}
\vskip -0.1in
\end{table}

\begin{figure}[hbt!] 
\centering
\subcaptionbox{Dependent Experts \label{fig:pumadyn_time_d}}{\includegraphics[width=0.45\columnwidth]{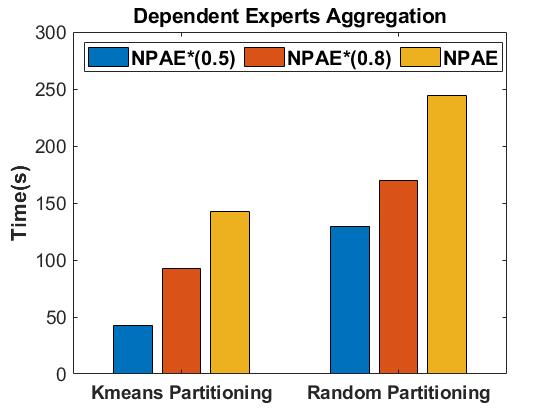}}\hfill
\subcaptionbox{Conditionally Independent Experts\label{fig:pumadyn_time_CI}}{\includegraphics[width=0.45\columnwidth]{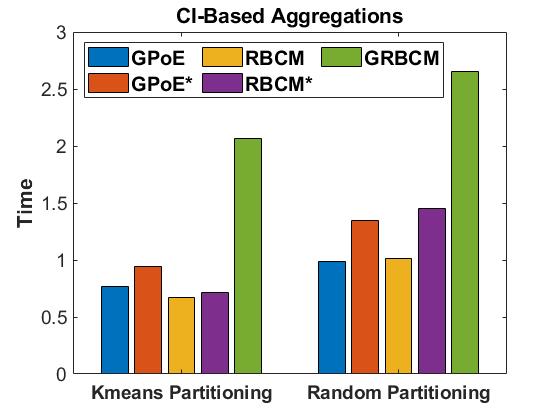}}\hfill
\caption{\textbf{Prediction Time} (seconds) of different aggregation for disjoint and random partitioning strategies in \textit{Pumadyn} data set. In both partitioning strategies, the training data set is divided into 15 subsets and a fixed penalty term ($\lambda=0.1$) is used.}
\label{fig.pumadyn_time}
\vskip -0.1in
\end{figure} 

Figure \ref{fig.pumadyn_time} depicts the prediction time of different baselines on the \textit{Pumadyn} data set using both partitioning strategies. Figure \ref{fig:pumadyn_time_d} compares the impact of our expert selection approach with NPAE and confirms that NPAE* is an appropriate fast approximation of NPAE. Its running time is up to one third that of NPAE while providing competitive results. On the other hand, Figure \ref{fig:pumadyn_time_CI} shows the computational costs of the CI-based methods and our suggested variants (GPOE*, RBCM*): the computational costs increase marginally while our modified versions improve upon prediction quality as shown in Table \ref{table.pumadyn}. 

\subsubsection{Large Data sets} \label{sec:4.2.2}
In this section we use three large-scale data sets, \textit{Protein}\footnote{\url{https://archive.ics.uci.edu/ml/datasets/Physicochemical+Properties+of+Protein+Tertiary+Structure}}, \textit{Sacros}\footnote{\url{http://www.gaussianprocess.org/gpml/data/}}, and \textit{Song}\footnote{\url{https://archive.ics.uci.edu/ml/datasets/yearpredictionmsd}}. \textit{Protein} has 9 dimensions and 45730 observations, $90\%$ of which are used for training (i.e. 41,157 training and 4,573 test points). \textit{Sacros} has 21 dimensions and 44,484 training and 4,449 test points. Finally, \textit{Song} is a 91-dimensional data set with 515,345 instances, divided into 463,715 training and 51,630 test examples. We extract the first $10^5$ songs from this data set for training and keep the original set of 51630 songs for testing. We used disjoint partitioning to divide the data sets into 70 (for \textit{Protein}), 72 (for \textit{Sacros}), and 150 (for \textit{Song}) subsets.

Next, we compare SOTA baselines with NPAE*. We also consider GPoE* and RBCM*.\footnote{In Appendix \ref{appendix:large_scale} we show the influence of the expert selection approach on the other CI-based models.} For the model selection based methods, we set $\alpha$ to $0.8$ which means $20\%$ of weak experts are excluded. For the \textit{Song} data set only, we set $\alpha=0.1$ in NPAE*.
Since NPAE is computationally burdensome, especially when $M$ and $n_t$ are large, it is not used in this part. 
\begin{table*}[hbt!]
\caption{\textbf{Prediction quality} for various methods on \textit{Protein}, \textit{Sacros}, and \textit{Song} data. The table depicts SMSE and MSLL values for SOTA  baselines and the modified versions of GPoE, RBCM, and NPAE, i.e. GPoE*, RBCM*, and NPAE* respectively.}
\label{table.kmeans}
\vskip -5 mm
\footnotesize
\begin{center}
\begin{small}
\begin{sc}
\resizebox{\textwidth}{!}{%
\begin{tabular}{lcccccccr}
\toprule
\multirow{2}{*}{} &
\multicolumn{1}{c}{} &
\multicolumn{2}{c}{Protein} &
\multicolumn{2}{c}{Sacros} & 
\multicolumn{2}{c}{Song} & \\
\toprule
Model & Type  & SMSE  & MSLL& SMSE  & MSLL & SMSE  & MSLL  \\
\midrule
PoE   & CI & 0.8553 &33.2404& 0.045 & 2.724 &  0.952 & 70.325 \\
GPoE  & CI & 0.8553 & -0.082 & 0.045 & -1.183 &  0.952& -0.029 \\
\textcolor{blue}{GPoE*}  & CI & 0.8146 & -0.1146 & 0.0038 & -1.282 & 0.943 & -0.034  \\
BCM    & CI & 0.3315 & -0.3329& 0.007 & -2.359 & 2.660 & 3.438 \\
RBCM   & CI & 0.3457& -0.584 & 0.0045 & -2.413 & 0.847 & -0.006 \\
\textcolor{blue}{RBCM*}  & CI &  0.3411 & -0.5953 & 0.0041 & -2.520 & 0.842 & -0.022 \\ 
GRBCM  & CI & 0.3477& -0.613& 0.0037 & -2.642 & 0.836 & -0.0926  \\
\midrule
\textcolor{blue}{NPAE*}  & D & \textbf{0.3101} & \textbf{-0.6653}  & \textbf{0.0028} & \textbf{-2.772} & \textbf{0.794} & \textbf{-0.110}\\
\bottomrule
\end{tabular}}
\end{sc}
\end{small}
\end{center}
\vskip -0.2in
\end{table*}

Table \ref{table.kmeans} reveals the prediction quality of local approximation baselines and shows NPAE* outperforms other SOTA methods on various data sets. In particular,  NPAE* provides significantly better predictions on the \textit{Song} data set using only 10 percent of mostly interacted experts. For CI-based methods, GPoE* and RBCM* improve the prediction quality of GPoE and RBCM method, respectively. On the \textit{Protein} data set, BCM and RBCM* have smaller SMSE values while the MSLL values of RBCM* and GRBCM are of the same rate. On the \textit{Sacros} data set, the SMSE values of GPoE*, RBCM*, and GRBCM are smaller than the other CI-based methods while RBCM* and GRBCM have smaller MSLL values. Although all CI-based methods return poor predictions on the \textit{Song} data set, RBCM* and GRBCM provide smaller SMSE while the MSLL value in GRBCM is smaller than other CI-based SOTA methods. Although the SMSE of RBCM and RBCM* are of the same rate, RBCM* significantly improves the MSLL values of the RBCM method.

\section{Conclusion}
In this work, we have proposed a novel expert selection approach for distributed GPs which leverages expert selection to aggregate dependent local experts' predictions. To combine correlated experts, comparable SOTA methods use all experts and are affected by weak experts or leading to impractically high computational costs. Our proposed approach uses an undirected graphical model to find the most important experts for the final aggregation. Theoretically, we showed that our new local approximation approach provides consistent results when $n \to \infty$. Through empirical analyses, we illustrated the superiority of our approach which improves the prediction quality of existing SOTA aggregation methods, while being highly efficient.

\bibliography{bibliography.bib}
\bibliographystyle{unsrt}

%%%%%%%%%%%%%%%%%%%%%%%%%%%%%%%%%%%%%%%%%%%%%%%%%%%%%%%%%%%%

\newpage
\appendix

\section{Expert Weighting for Distributed GP Models}\label{sec:DGP}
In this section we review other divide-and-conquer distributed GP approaches in more detail, focusing on how other methods perform expert weighting. There are two main families of distributed GPs: product of experts (PoE) and Bayesian committee machine (BCM). 
\paragraph{Product of Experts.}
The posterior distribution of the PoE model is given by the product of multiple densities (i.e., the experts). Because of the product operation, the prediction quality of PoE suffers considerably from weak experts. To improve on this aspect, \cite{Cao} proposed the GPoE model, which assigns importance weight to the experts.

For independent experts $\{\mathcal{M}\}_{i=1}^M$ trained on different partitions $\mathcal{D}_i$ the predictive distribution for a test input $X^*$ is given by:
\begin{equation} \label{eq:13}
p(y^*|\mathcal{D},X^*)= \prod_{i=1}^M p_i^{\beta_i}(y^*|\mathcal{D}_i,X^*),
\end{equation}
where $\beta = \{\beta_1,\ldots,\beta_M\}$ controls the expert importance. The product distribution in \eqref{eq:13} is proportional to a Gaussian distribution with mean and precision, respectively:
\begin{align} 
\mu_{D}^* &= \Sigma_{D}^* \sum_{i=1}^M \beta_i(\Sigma_i^*)^{-1}\mu_i^*, \label{eq:14} \\ 
(\Sigma_{D}^*)^{-1} &= \sum_{i=1}^M \beta_i (\Sigma_i^*)^{-1}.  \label{eq:15}
\end{align}

The standard PoE can be recovered by setting $\beta_i=1 ~ \forall i$. The precision corresponding to the PoE prediction in \eqref{eq:15} is a linear sum of individual precision values: hence, an increasing number of local GPs increases the precision and therefore it leads to a decrease in variance, which consequently returns overconfident predictions in areas with little data.

To choose the weights $\beta_i$ in the PoE model several heuristics have been put forward. The authors of \cite{Cao} suggested the difference in differential entropy between the prior and posterior distribution of each expert, i.e. $\beta_i=\frac{1}{2}(\log \Sigma^{**} - \log \Sigma_i^{*})$ where the $(\Sigma^{**})^{-1}$ is the prior precision of $p(y^*)$. This leads to more conservative predictions. To fix this issue, \cite{Deisenroth} suggested to choose simple uniform weights $\beta_i=\frac{1}{M}$, which provides better predictions.
\paragraph{Bayesian Committee Machine.}
The Bayesian committee machine (BCM) \cite{Tresp} uses the Gaussian process prior $p(y^*)$ for the aggregation step and assumes conditional independence between experts, i.e. $\mathcal{D}_i \perp \!\!\! \perp  \mathcal{D}_j |y^*$ for two experts $i$ and $j$. 
To mitigate the effect of weak experts on aggregation, especially in regions with few data points, \cite{Deisenroth}
proposed the robust Bayesian committee machine (RBCM), which added importance weights $\beta_i$ to the BCM model.
The distributed predictive distribution of this family of models can be written as: 
\begin{equation*} %\label{eq:16}
p(y^*|\mathcal{D},X^*)= \frac{\prod_{i=1}^M p_i^{\beta_i}(y^*|\mathcal{D}_i,X^*)}{p^{\sum_{i=1}^M \beta_i-1}(y^*)}.
\end{equation*}
Its distribution is proportional to a Gaussian distribution with mean and precision, respectively:
\begin{align*} 
\mu_{D}^* &= \Sigma_{D}^* \sum_{i=1}^M \beta_i(\Sigma_i^*)^{-1}\mu_i^*, \\ %\label{eq:17} 
(\Sigma_{D}^*)^{-1} &= \sum_{i=1}^M \beta_i (\Sigma_i^*)^{-1} +(1-\sum_{i=1}^M \beta_i)(\Sigma^{**})^{-1},  %\label{eq:18}
\end{align*}
where the $(\Sigma^{**})^{-1}$ is the prior precision of $p(y^*)$. The general choice of the weights is the difference in differential entropy between the prior $p(y^*|X^*)$ and the posterior $p(y^*|\mathcal{D},X^*)$, i.e. $\beta_i=\frac{1}{2}(\log \Sigma^{**} - \log \Sigma_i^{*})$.

The most recent model in the BCM family is the generalized robust Bayesian committee machine (GRBCM) \cite{Liu2018}. It introduces a base (global) expert and considers the covariance between the base and other local experts. For a global expert $M_b$ and a base partition ${D}_{b}$, the predictive distribution of GRBCM is
\begin{equation}
p(y^*|\mathcal{D},X^*)= \frac{\prod_{i=2}^M p_{bi}^{\beta_i}(y^*|\mathcal{D}_{bi},X^*)}{p_b^{\sum_{i=2}^M \beta_i-1}(y^*|\mathcal{D}_b,X^*)} ,
\end{equation}
where the $p_b(y^*|\mathcal{D}_b,X^*)$ is the predictive distribution of $M_b$, and $p_{bi}(y^*|\mathcal{D}_{bi},X^*)$ is the predictive distribution of an expert trained on the data set $\mathcal{D}_{bi}=\{\mathcal{D}_{b},\mathcal{D}_{i}\}$. The base partition is randomly selected, while the remaining experts can be chosen through a random or disjoint partitioning strategy. It is noteworthy that, for $M$ experts and $m_0$ data points per expert, the GRBCM operates based on $M-1$ experts with $2m_0$ data points per expert. Therein lies the main difference between GRBCM and the other distributed GPs, which use $m_0$ data points per expert only. Since GRBCM assigns more data points to the experts, it trains experts on more informative subsets. %Therefore, it provides almost confident predictions, especially for disjoint data partitioning regimes. %The prediction process of GRBCM has time complexity $\mathcal{O}(\alpha nm^2_0) + \mathcal{O}(\beta n'n m_0)$, where $m_0$ in the number of assigned points to each expert, $n'$ is the size of test set, $\alpha=(8M-7)/M$, and $\beta=(4M-3)/M$. 

\section{On the relation between Experts Weights and expert selection approach.} \label{experts_weights}
Assume a test point $x^*$ is far away from a data partition $\mathcal{D}_i$. Using this partition, the prediction will generally be poor. Then, $k(x^*,X_i) \approx 0$, and $\Sigma_i^{*} \approx \Sigma^{**}$. Suppose the weights are defined as the difference in differential entropy. Then $\beta_i \approx 0$. Thus CIA based distributed GPs tend to assign a small weight to $\mathcal{D}_i$. On the other hand, in our expert selection approach, since  $\mathcal{D}_i$ provides a poor prediction at $x^*$, its interactions with the other experts are weak, see \eqref{eq:5} when $\Gamma_i\approx 0$. Hence, our approach recognizes this expert as an irrelevant expert. Consequently, expert selection eliminates this expert, while CIA based distributed GPs tend to keep this expert and assign a small weight to $\mathcal{D}_i$.

Finally, if equal weights are used for the CIA based methods, i.e. $\beta_i=1/M$, weak and strong experts have an equal effect on the aggregated predictions. This may not be appropriate, particularly when there is a large number of candidate experts.

\section{Proof of Proposition \ref{prop:predictive_distribution}}\label{sec:proof}
The proof of part (I) and (II) is obvious using the property of conditional Gaussian distribution and NPAE method. We here proof part (III). 

Our proof is based on disjoint $K$-means partitioning, and the proof for random partitioning is similar. For the proof, we need to show that the importance-based estimator $y_{\alpha}^*(x^*)$ is almost equivalent to $y_A^*(x^*)$ in \eqref{aggregate_mean}. To do that, we partition both $k_A(x^*) \text{ and } K_A(x^*)^{-1}$ with respect to $\tilde{\mathcal{M}}_{\alpha}$ and $\tilde{\mathcal{M}}^c_{\alpha}$. Hence, we obtain $k_A(x^*)=\left[k_{\alpha}(x^*),k_{c}(x^*)\right]$, $\mu^*(x^*)=\left[\tilde{\mu}_{\alpha}^*(x^*),\mu^*_{c}(x^*)\right]$, and 
$$
K_A(x^*)^{-1} =
\begin{bmatrix}
K_{\alpha}(x^*)^{-1} & K_{\alpha c}(x^*)^{-1} \\
K^t_{\alpha c}(x^*)^{-1} & K_c(x^*)^{-1}
\end{bmatrix}.
$$
Due to the definition of $\tilde{\mathcal{M}}^c_{\alpha}$, we have $K_{\alpha c}(x^*)^{-1} \approx 0$. Note that $K_{c}(x^*)^{-1}$ is approximately a diagonal matrix, where the diagonal elements are pointwise variances of experts' predictions corresponding to \eqref{eq:3}. For the sake of better readability, we omit $x^*$ from our notation. Now, we have:
\begin{align*}
y_A^*&=k_A^T K_A^{-1}\mu^* =\left[k_{\alpha},k_{c}\right] 
\begin{bmatrix}
K_{\alpha}^{-1} & K_{\alpha c}^{-1} \\
K_{\alpha c}^{-t} & K_c^{-1}
\end{bmatrix} 
\left[\tilde{\mu}_{\alpha}^*,\mu^*_{c}\right]\\
&= k_{\alpha} K_{\alpha}^{-1}\tilde{\mu}_{\alpha}^* + k_{c} K_{\alpha c}^{-t} \tilde{\mu}_{\alpha}^* + k_{\alpha} K_{\alpha c}^{-1} \mu^*_{c} + k_{c} K_{c}^{-1} \mu^*_{c} \\
& \approx  k_{\alpha} K_{\alpha}^{-1}\tilde{\mu}_{\alpha}^* + k_{c} K_{c}^{-1} \mu^*_{c}.
\end{align*}
What remains to be shown is that when $n \to \infty$, $ k_{c} K_{c}^{-1} \mu^*_{c}\approx 0$. Let $\mathcal{M}_{c_i}$ be the $i$-th partition of $\tilde{\mathcal{M}}^c_{\alpha}$. Since $\mathcal{M}_{c_i}$ does not have strong interactions with other experts given $x^*$, the off-diagonal values of $K_A(x^*)^{-1}$ related to this partition are small. Since the interactions between experts are estimated using their local predictions, small interactions between this partition and the others means its prediction $\mu^*_{c_i}$ is not close to the other experts' predictions.\footnote{We need to mention that at each test point, the correlation between two partitions (or experts)is the correlation between their local predictions which depends on their observations and also the test point, see \eqref{eq:5} and the definition of $\Gamma_i$.} According to the disjoint partitioning method, we can conclude that $x^*$ is away from the training partition $\mathcal{M}_{c_i}$, see Section \ref{experts_weights} for more details\footnote{If $x^*$ is close to the training partition $\mathcal{M}_{c_i}$, then its local predictions should be approximately close to the local predictions of its neighbor partitions and therefore, it should have strong interactions with them but it not true due to the definition of $\tilde{\mathcal{M}}^c_{\alpha}$.}. It is clear that the further $x^*$ is away from a partition, the higher the relative distance $r_{c_i}=\lVert x^* - x \rVert_{x\in\mathcal{M}_{c_i}}$ grows. Using assumption (ii) stated in proposition \ref{prop:predictive_distribution}, we have $r_{c_i} \to \infty$ since $n \to \infty$.\footnote{The term $r_{c_i} \to \infty$ when $n \to \infty$ says: by increasing the training data sample size, the distance between a test point, and the partitions, that are away from the test point, increases.} Thus we have $\lim_{r_{c_i} \to \infty} \Sigma_i^* = k_{**}$, i.e. $k(x^*,X_{c_i})\to 0$. Since $\mu^*$ contains the centered experts' predictions, W.L.O.G assume that $\delta$ is an upper bound for the values of $\mu^*_c$. Then
\begin{align*}
(k_{c} K_{c}^{-1} \mu^*_{c})_i &\leq	k(x^*, X_{c_i}) \left[k(x*,x^*) - k(x^*,X_{c_i})^T (K_{c_i}+\sigma^2I)^{-1} k(x^*,X_{c_i})\right]^{-1} \delta \to 0.
\end{align*}
Therefore, $y_{A}^*(x^*) \approx y_{\alpha}^*(x^*)$ when $n \to \infty$. As a direct consequence, $y_{\alpha}^*(x^*)$ inherits the asymptotic properties of $y_{A}^*(x^*)$. A detailed proof, showing that $y_{A}^*(x^*)$ is statistically consistent can be found in \cite{Bachoc} (see their proof of Proposition 2), which employs a triangular array of observations. 

\section{Computational Costs}\label{sec:Computational_Costs}
The expert selection step significantly reduces the prediction cost of the NPAE method. The prediction cost of the original NPAE method is $\mathcal{O}(n_t M^3)$, where $n_t$ is the number of available test points and M is the number of experts. Glasso has also cubic time complexity $\mathcal{O}(M^3)$. Therefore, the complexity of NPAE* is approximately $\mathcal{O}(n_t M_{\alpha}^3 + M^3) \approx \mathcal{O}\left( (n_t \alpha^3 +1) M^3 \right)$ where $\alpha <1$. It means the cost of Glasso can be ignored when $n_t$ is large and the aggregation complexity of NPAE* is much lower than NPAE. In CI-based methods, the complexities of the original and modified versions are of the same rate when the number of experts is large, see Section \ref{sec:4}. 

\section{Extra Experiments about Important Experts }

In this section, we present some extra figures that show the most important experts and the experts with better individual prediction quality using synthetic data set in Section \ref{sec.4.1} and real-world data set in Section \ref{sec:4.2}. We use different number of experts and depicts the graphical models.

\subsection{Aggregation Based on Important and Unimportant Experts}
\begin{figure} [hbt!] 
\centering
\subcaptionbox{MAE (NPAE*)}{\includegraphics[width=0.3\columnwidth]{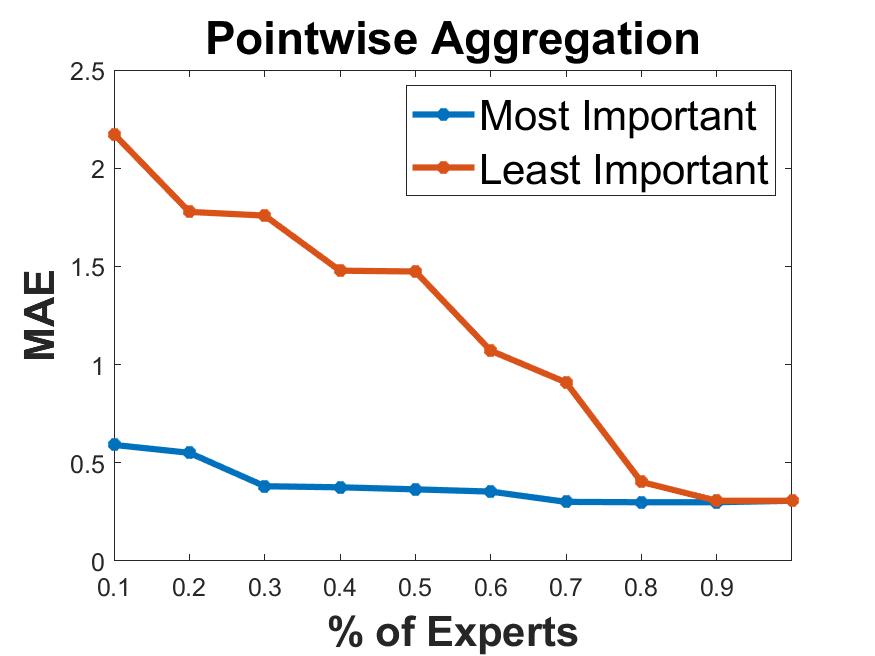}}
\subcaptionbox{SMSE (NPAE*)}{\includegraphics[width=0.3\columnwidth]{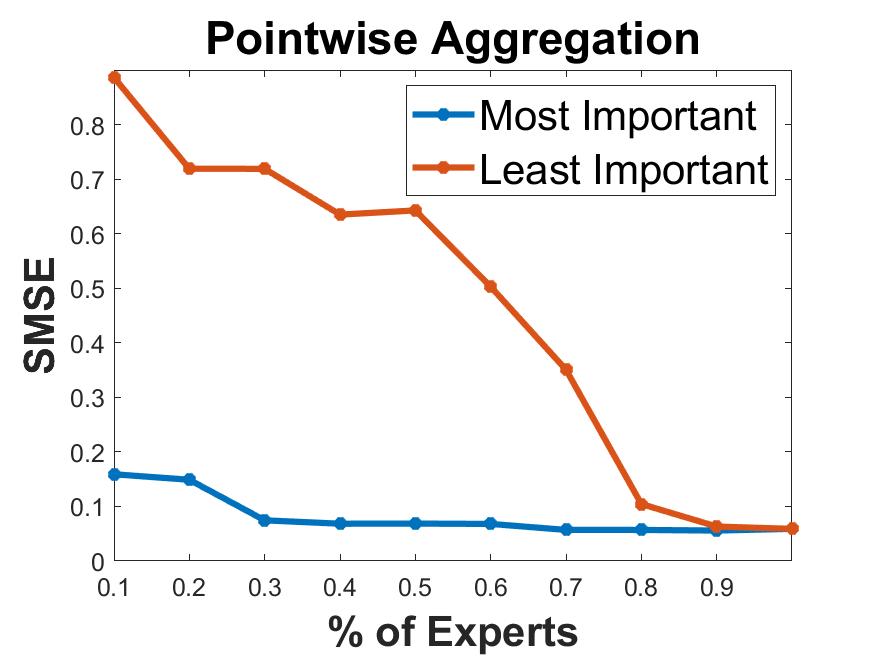}}
\subcaptionbox{MSLL (NPAE*)}{\includegraphics[width=0.3\columnwidth]{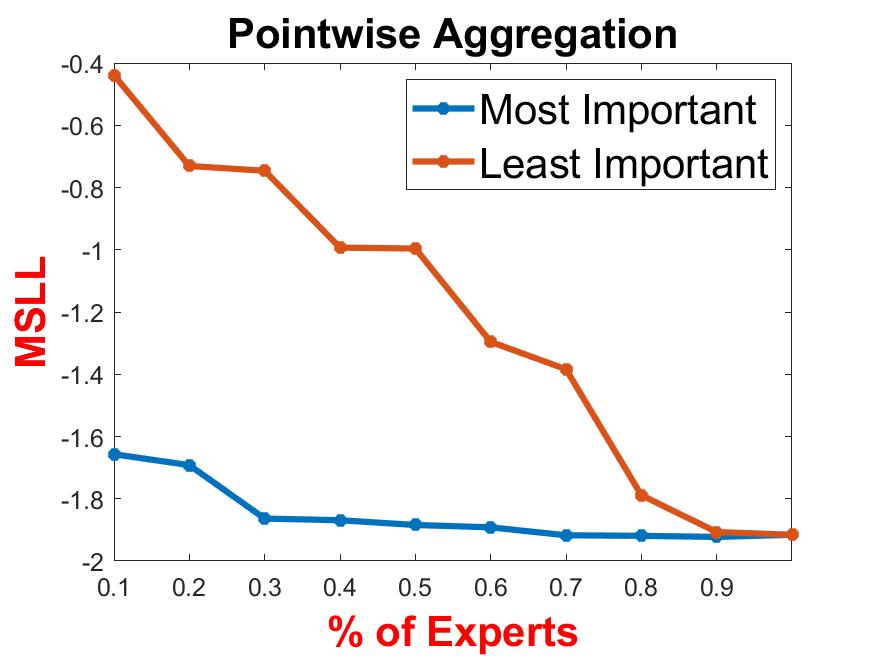}}
\caption{\textbf{Ablation experiment 2.} Expert selection for synthetic data from \eqref{f_x} with varying expert set strength $\alpha$ and $\lambda=0.1$. The blue line shows the prediction quality when $\alpha\%$ of most important experts are chosen using the GGM. The red line shows the prediction quality when the least important experts (according to Definition \ref{def.3}) are chosen.}
\label{fig.importance}
\vskip -0.1in
\end{figure}

Figure \ref{fig.importance} presents the effect of Definition \ref{def.3} on the prediction quality. The quality of predictions is evaluated in three ways: the mean absolute error (MAE), the standardized mean squared error (SMSE), and the mean standardized log loss (MSLL). We can see a significantly better quality of the aggregation based on the most important experts compared to the least important ones in the  NPAE* method. 

\subsection{Best Experts vs. Important Experts}
Similar to Figure \ref{fig.net}, here we depict the related GGMs of synthetic data set and \textit{Pumadyn} data set for different M and $\alpha$ values.
\begin{figure} [hbt!] 
\centering
\subcaptionbox{$M=10, \alpha=50\%$}{\frame{\includegraphics[width=0.32\columnwidth]{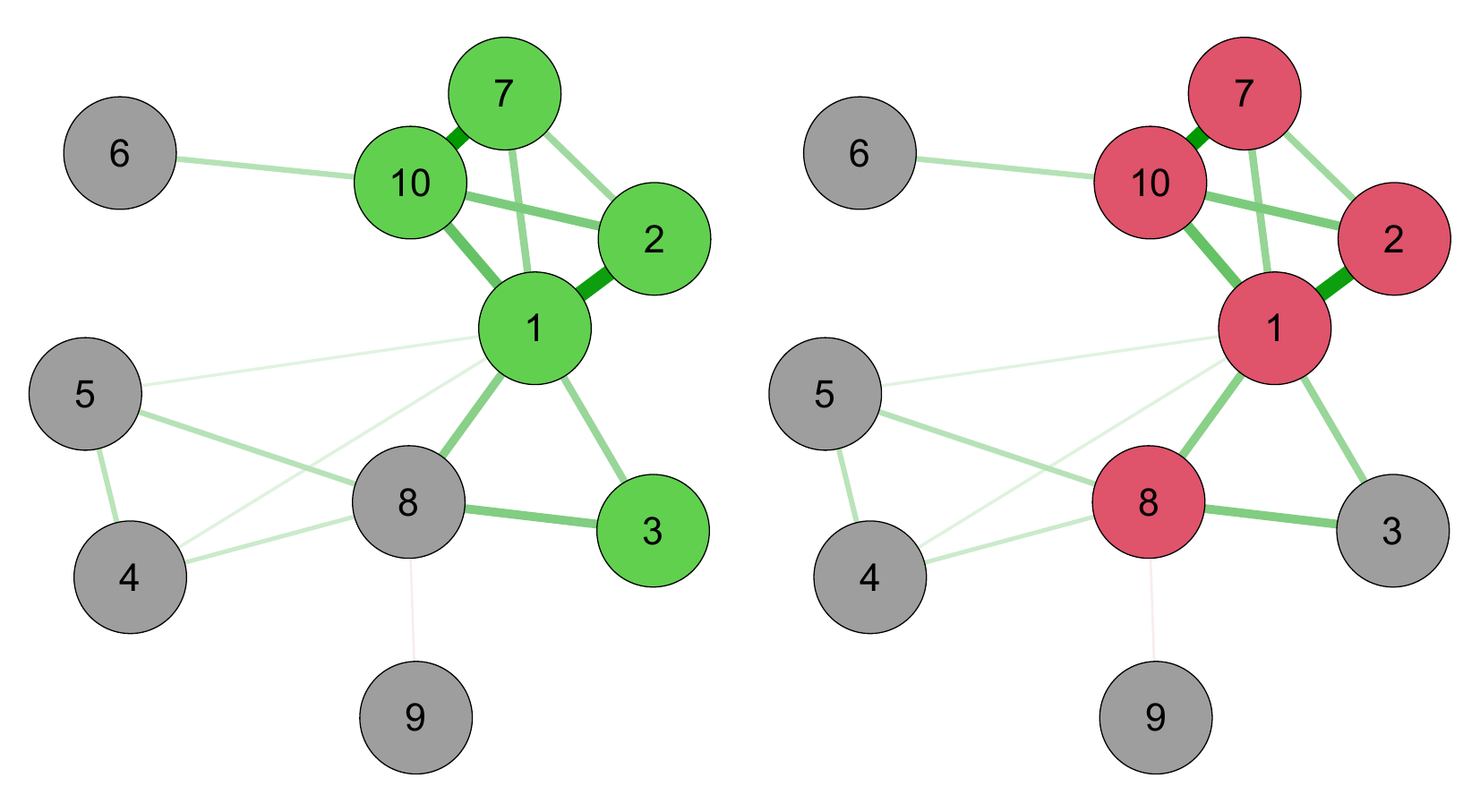}}}
\subcaptionbox{$M=10, \alpha=70\%$}{\frame{\includegraphics[width=0.32\columnwidth]{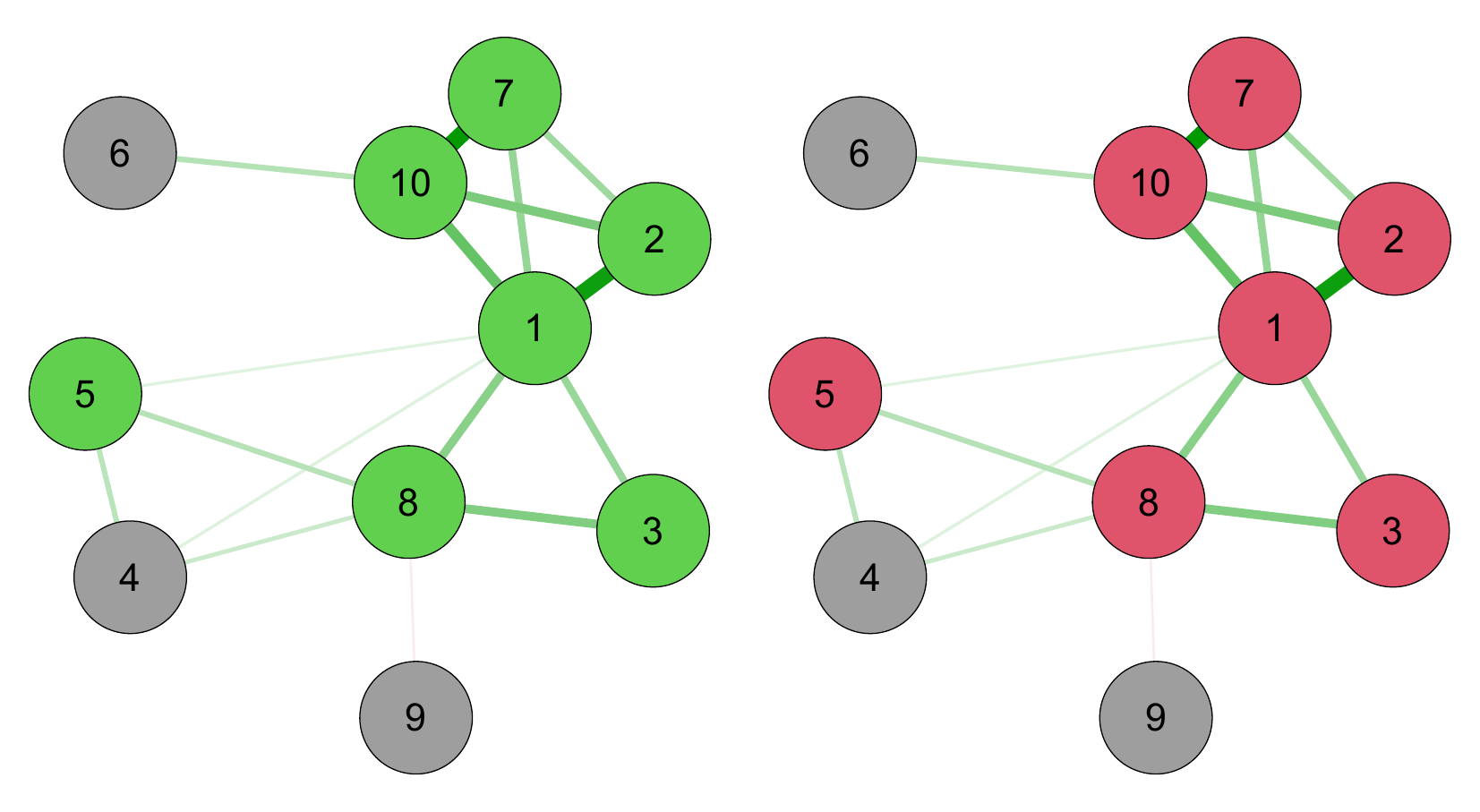}}}
\subcaptionbox{$M=10, \alpha=80\%$}{\frame{\includegraphics[width=0.32\columnwidth]{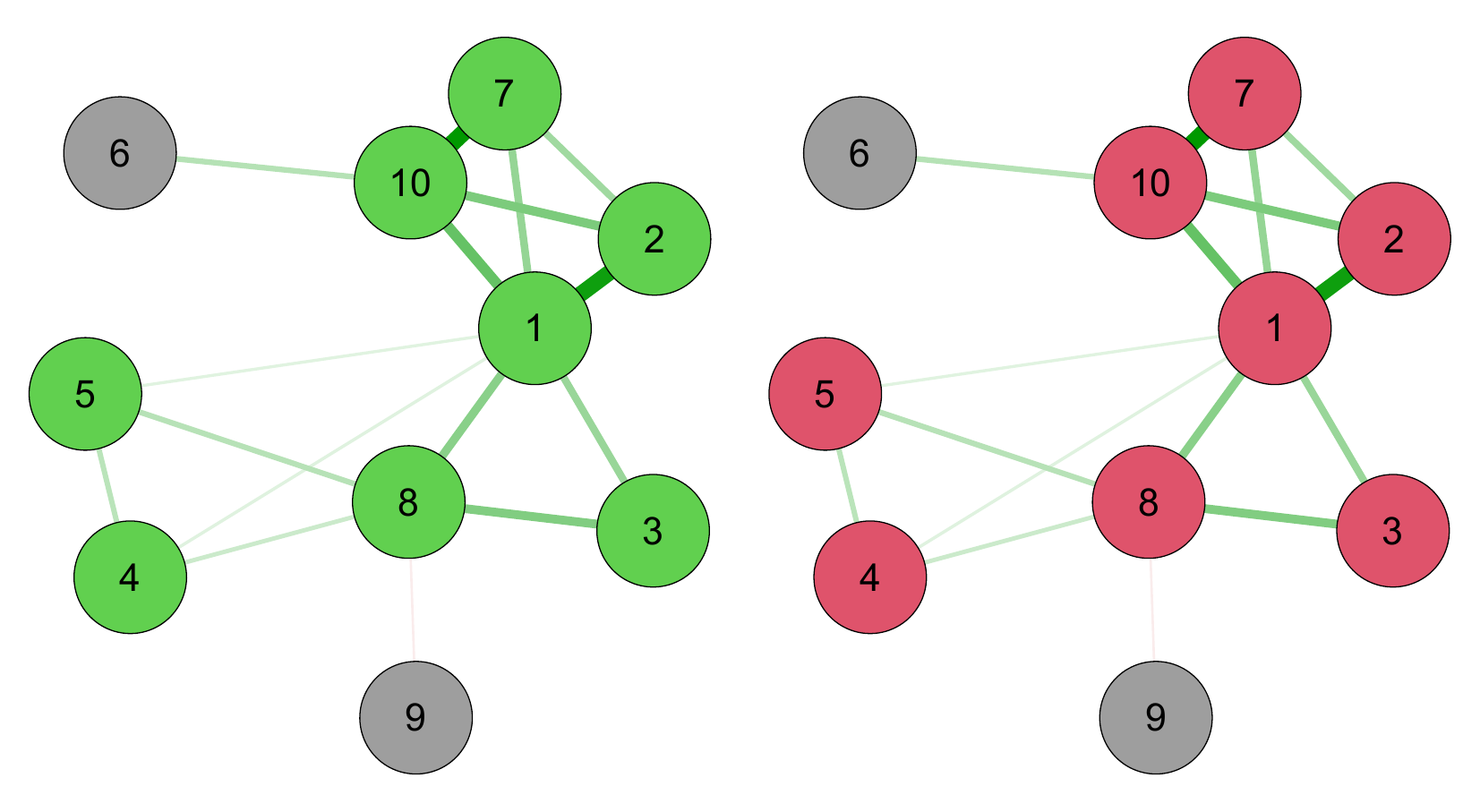}}}

\subcaptionbox{$M=15, \alpha=50\%$}{\frame{\includegraphics[width=0.32\columnwidth]{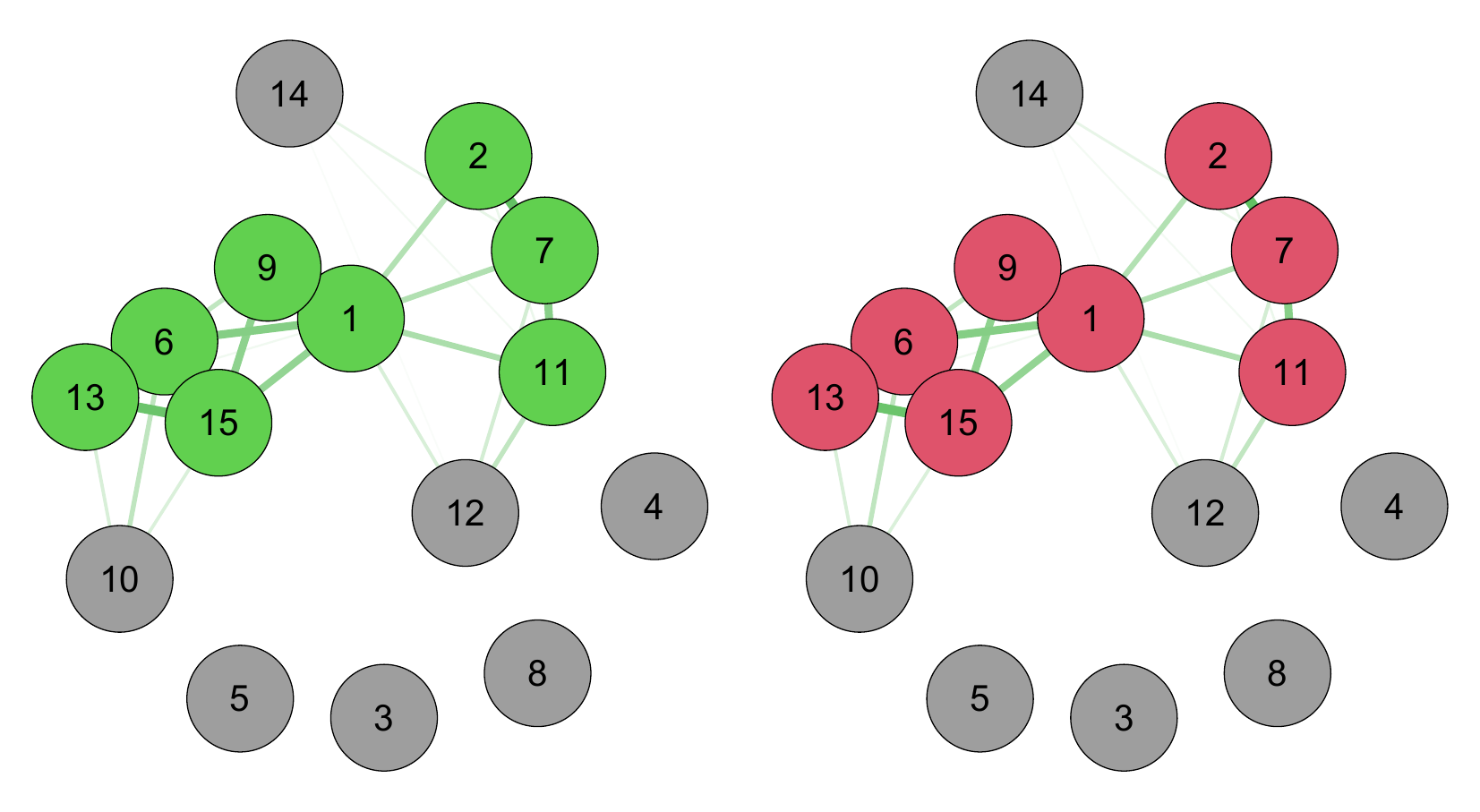}}}
\subcaptionbox{$M=15,\alpha=70\%$}{\frame{\includegraphics[width=0.32\columnwidth]{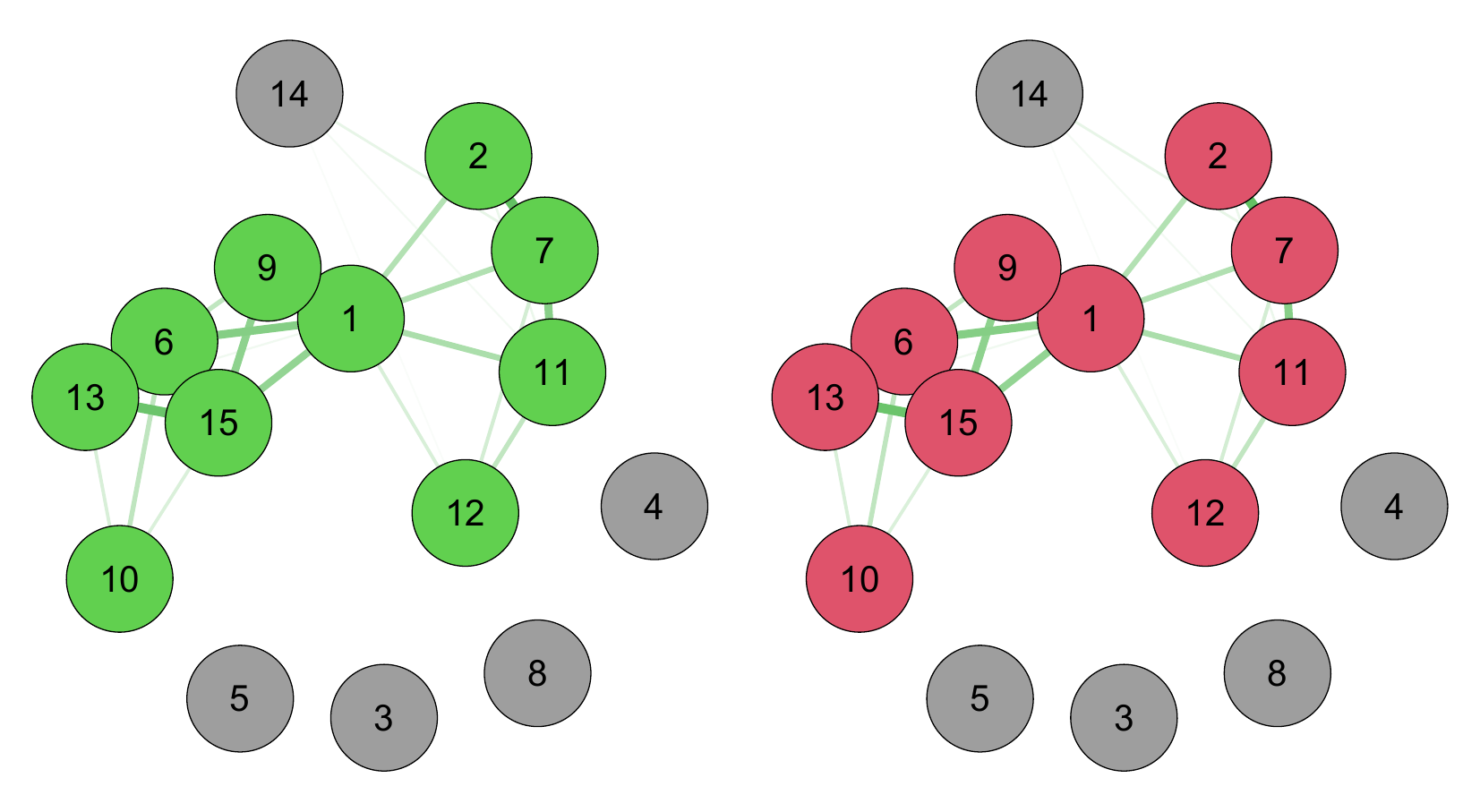}}}
\subcaptionbox{$M=15,\alpha=80\%$}{\frame{\includegraphics[width=0.32\columnwidth]{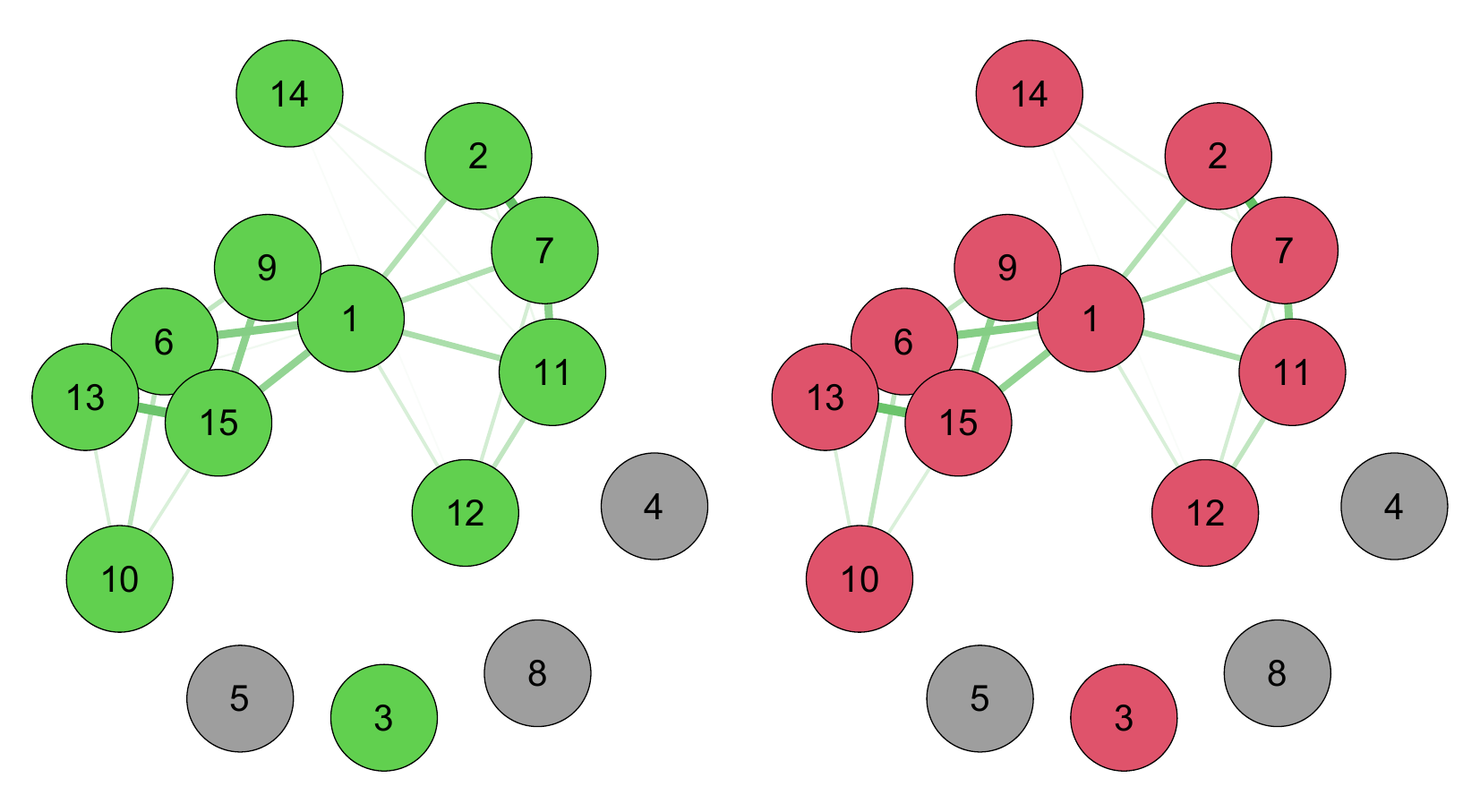}}}
%\subcaptionbox{$\alpha=75\%$}{\includegraphics[width=0.40\columnwidth]{net75.pdf}}
\caption{\textbf{Ablation experiment 3.} Expert selection for synthetic data from \eqref{f_x} with varying number of experts M, expert set strength $\alpha$, and $\lambda=0.1$. The green nodes reveal the $\alpha \%$ of the best best experts w.r.t. their individual MSE errors while the red nodes are the most important experts according to Definition \ref{def.3}.}
\label{fig.net_fx_2}
\vskip -0.1in
\end{figure}

\begin{figure} [hbt!] 
\centering
\subcaptionbox{$M=15, \alpha=50\%$}{\frame{\includegraphics[width=0.32\columnwidth]{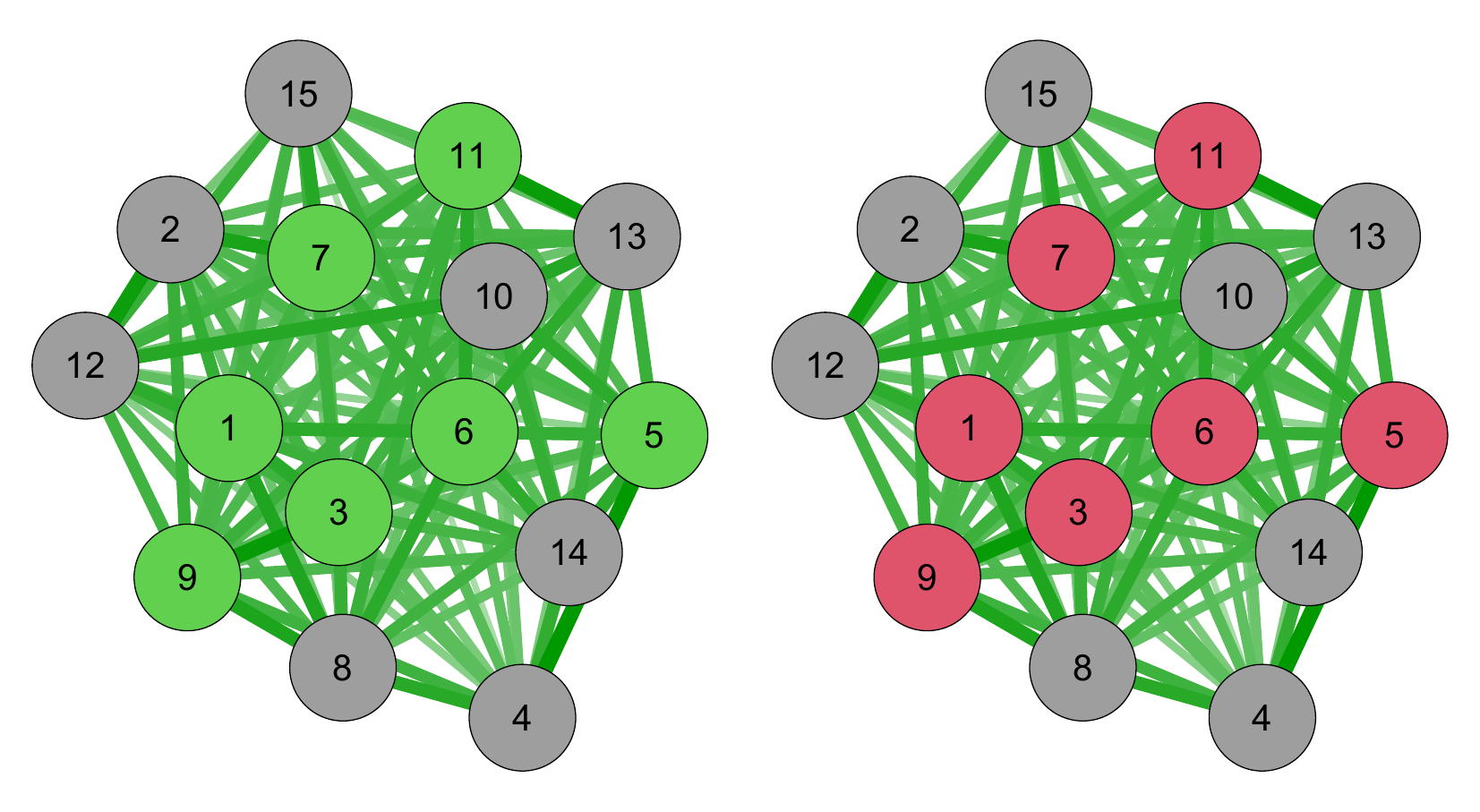}}}
\subcaptionbox{$M=15, \alpha=70\%$}{\frame{\includegraphics[width=0.32\columnwidth]{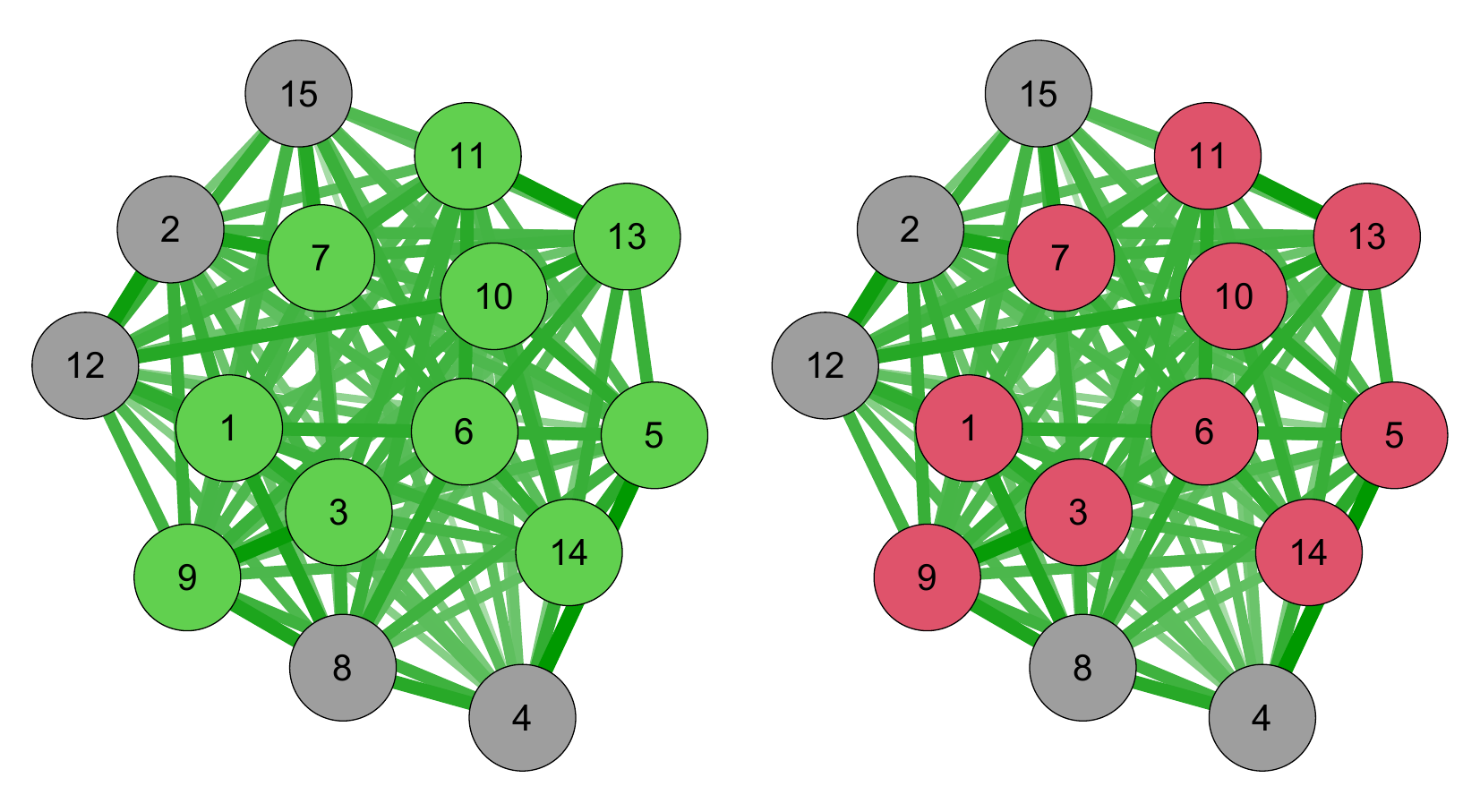}}}
\subcaptionbox{$M=15, \alpha=80\%$}{\frame{\includegraphics[width=0.32\columnwidth]{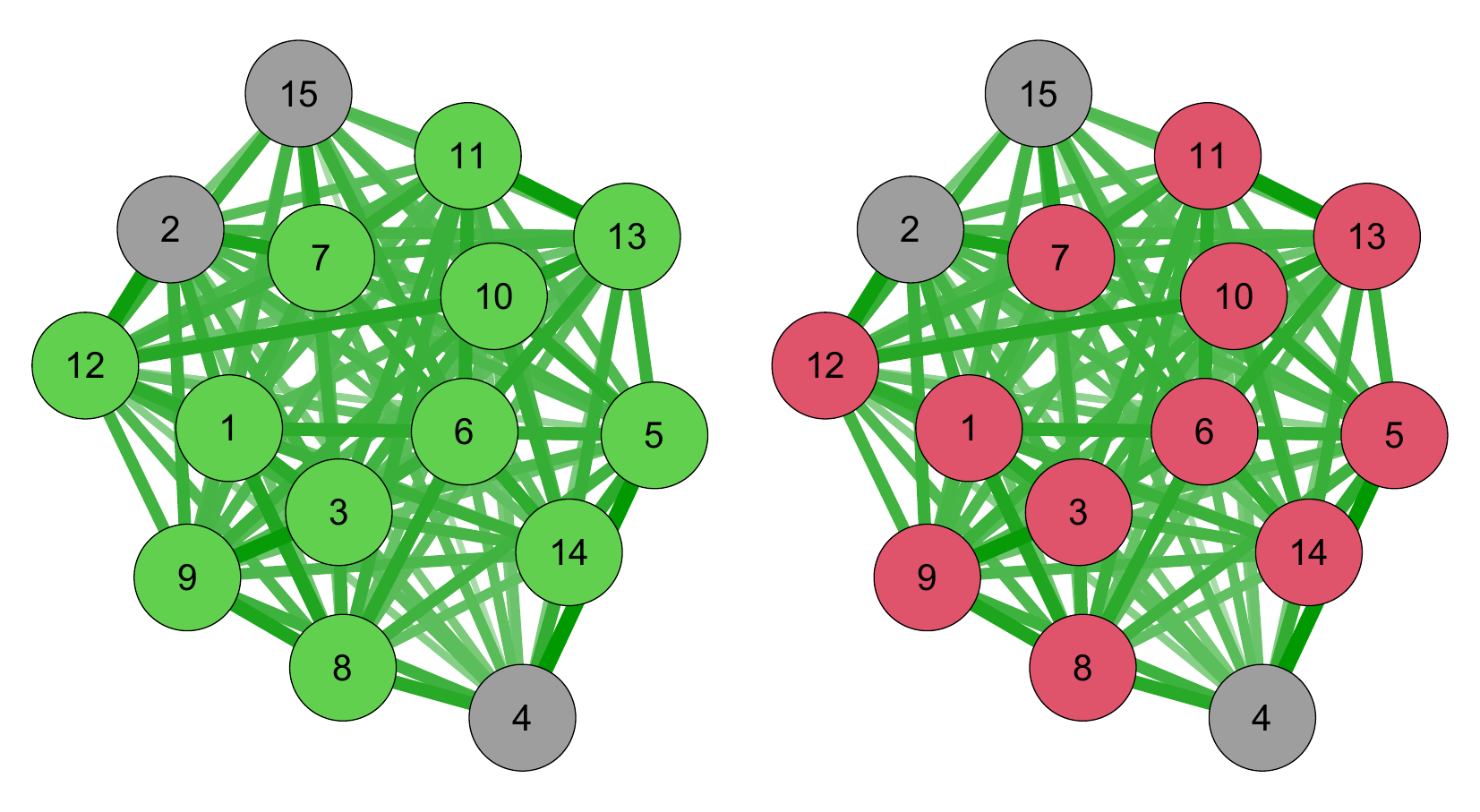}}}

\subcaptionbox{$M=20, \alpha=50\%$}{\frame{\includegraphics[width=0.32\columnwidth]{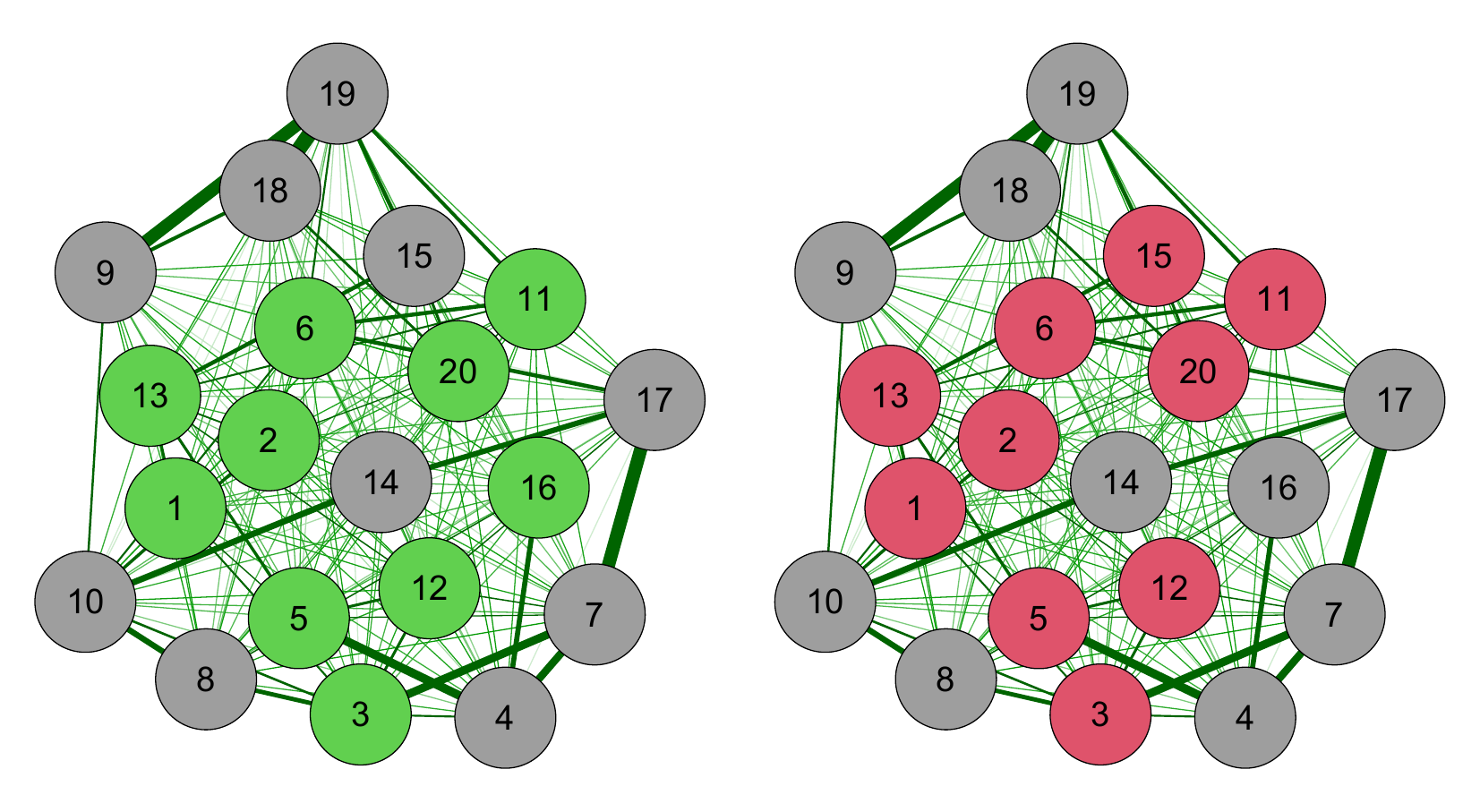}}}
\subcaptionbox{$M=20,\alpha=70\%$}{\frame{\includegraphics[width=0.32\columnwidth]{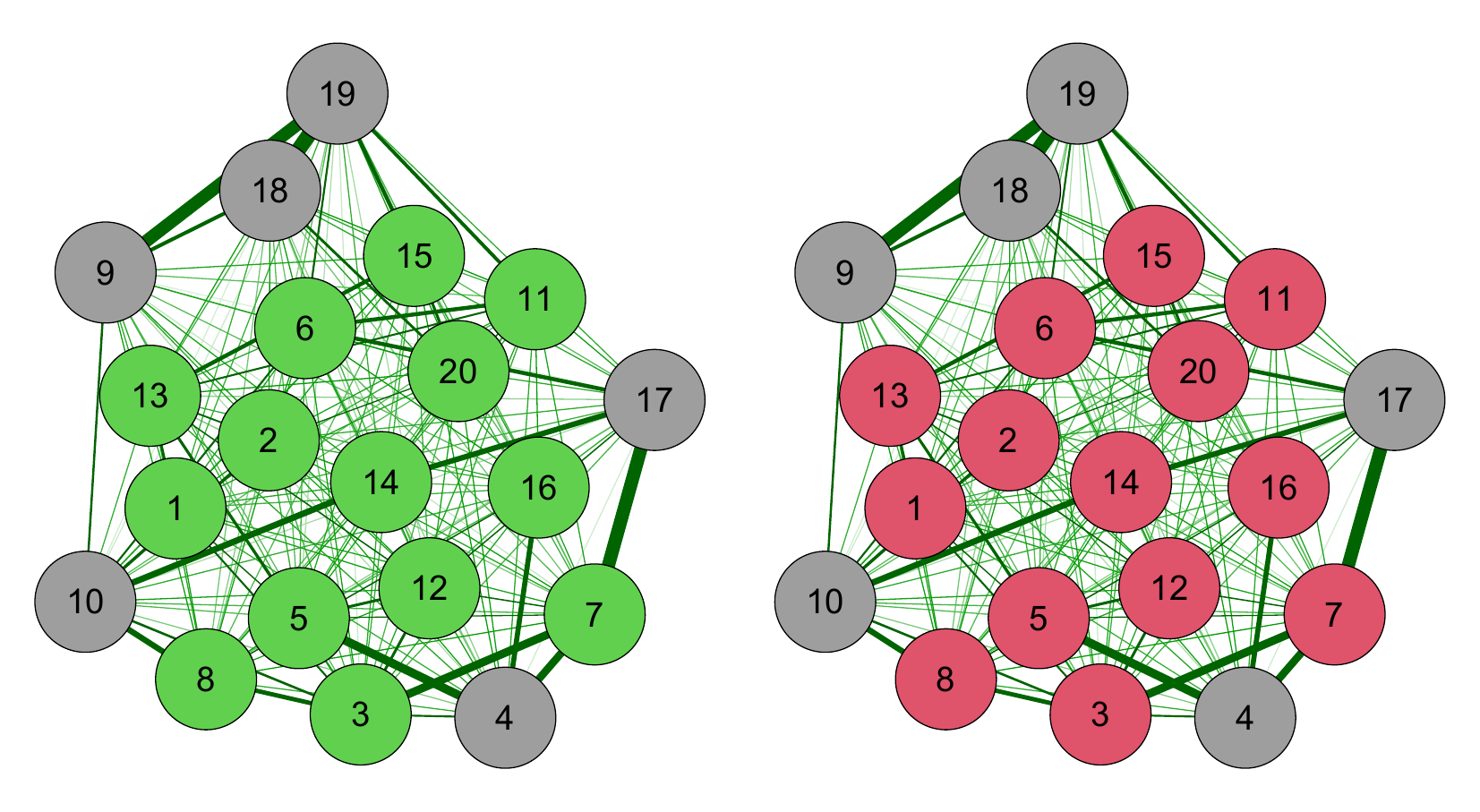}}}
\subcaptionbox{$M=20,\alpha=80\%$}{\frame{\includegraphics[width=0.32\columnwidth]{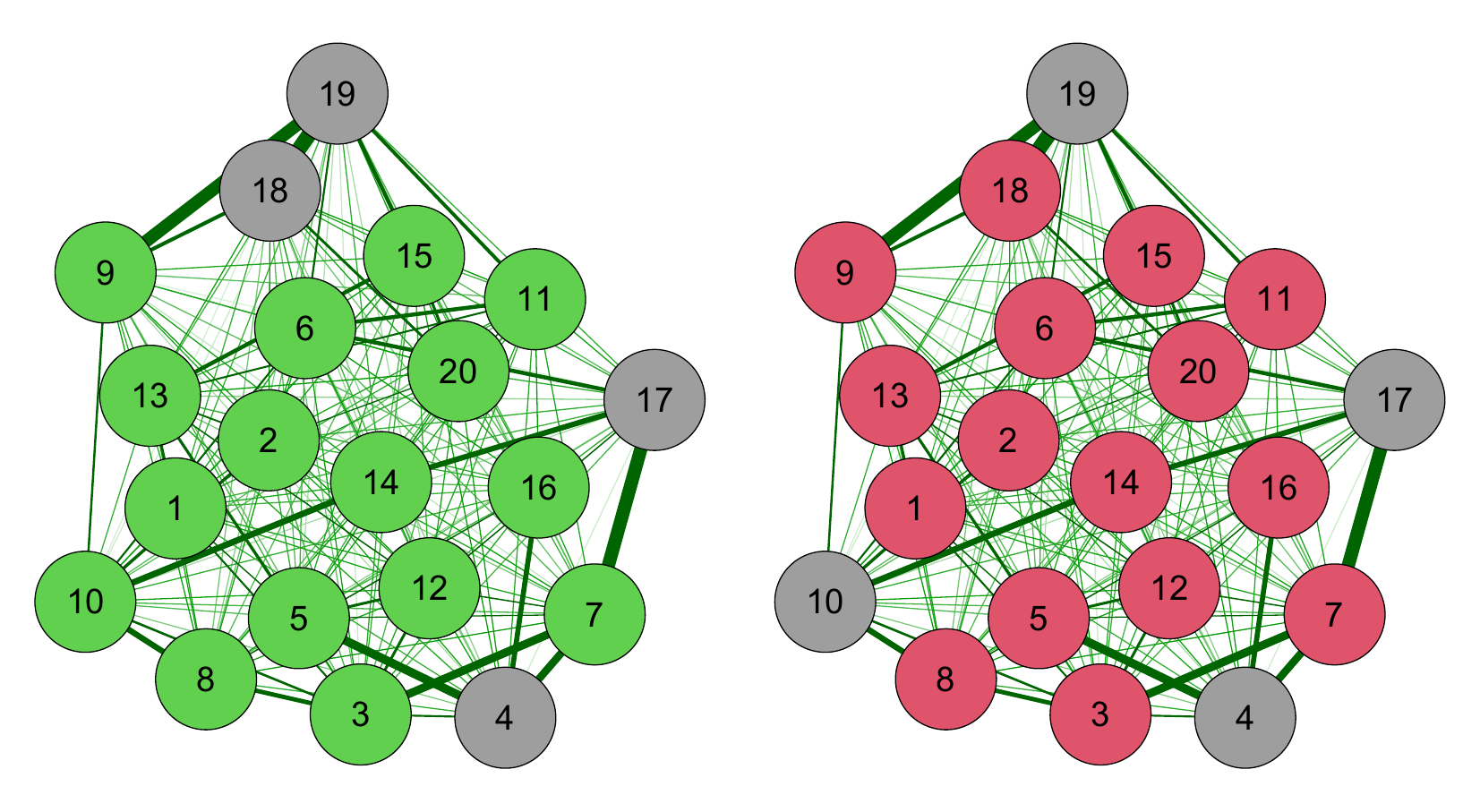}}}
%\subcaptionbox{$\alpha=75\%$}{\includegraphics[width=0.40\columnwidth]{net75.pdf}}
\caption{\textbf{Ablation experiment 4.} Expert selection for \textit{Pumadyn} data from with varying number of experts M, expert set strength $\alpha$, and $\lambda=0.1$. The green nodes reveal the $\alpha \%$ of the best best experts w.r.t. their individual MSE errors while the red nodes are the most important experts according to Definition \ref{def.3}.}
\label{fig.net_fx}
\vskip -0.1in
\end{figure}
%\newpage

\section{Experimental Results: Synthetic Example }\label{sec:experiments}
\begin{figure}[hbt!] 
\centering
\subcaptionbox{NPAE*(0.5) }{\includegraphics[width=0.33\columnwidth]{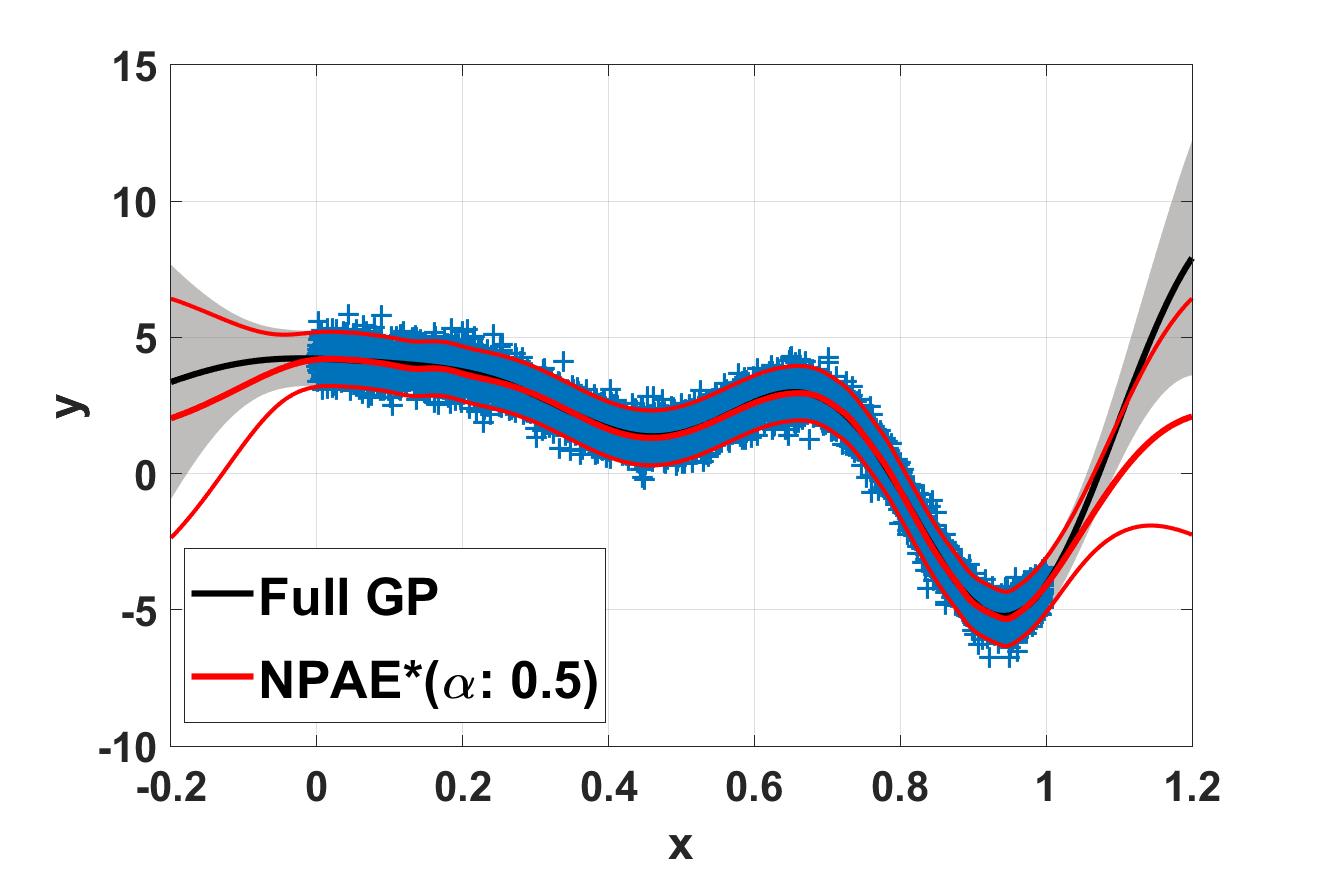}}\hfill
\subcaptionbox{NPAE*(0.8) }{\includegraphics[width=0.33\columnwidth]{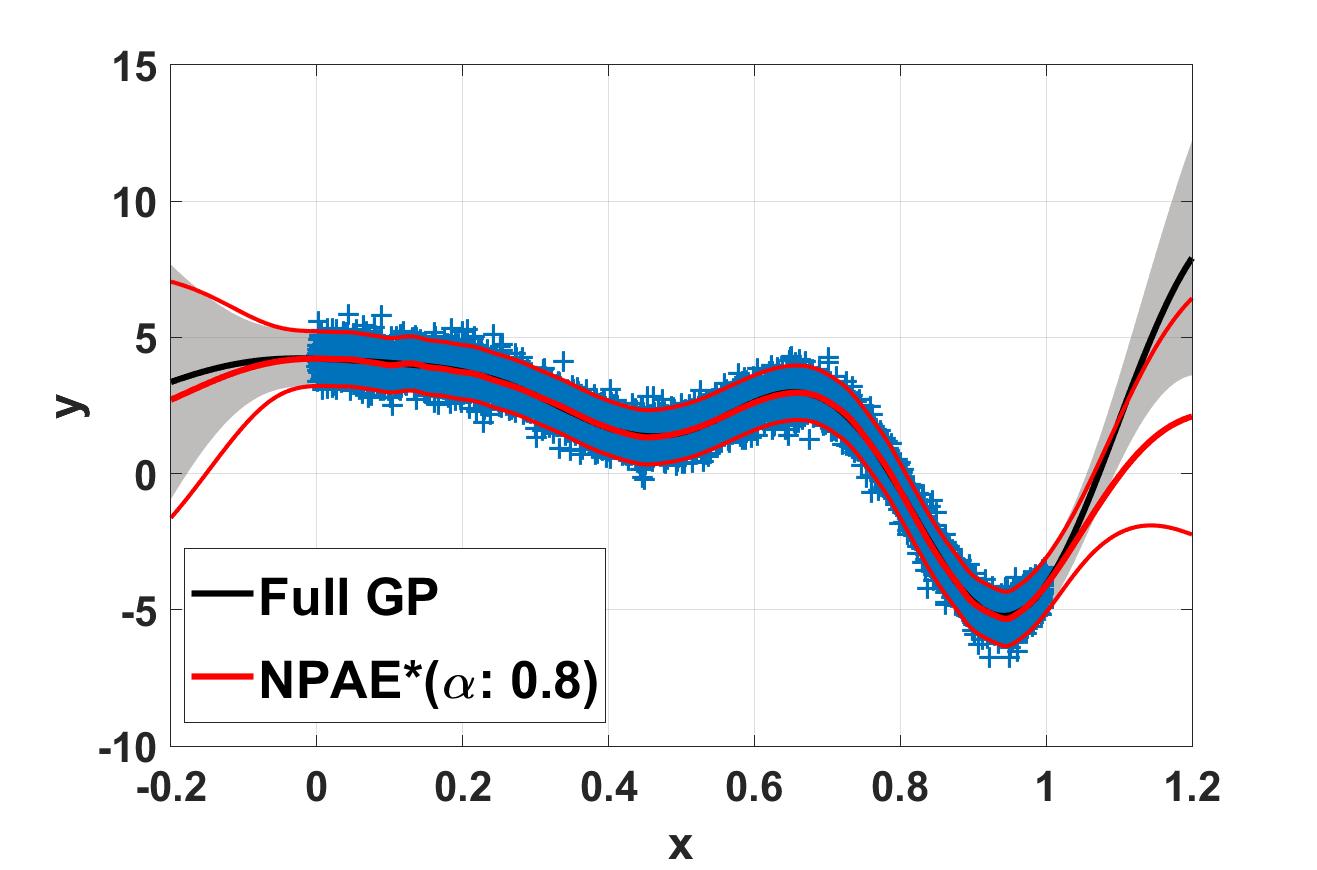}}\hfill
\subcaptionbox{NPAE }{\includegraphics[width=0.33\columnwidth]{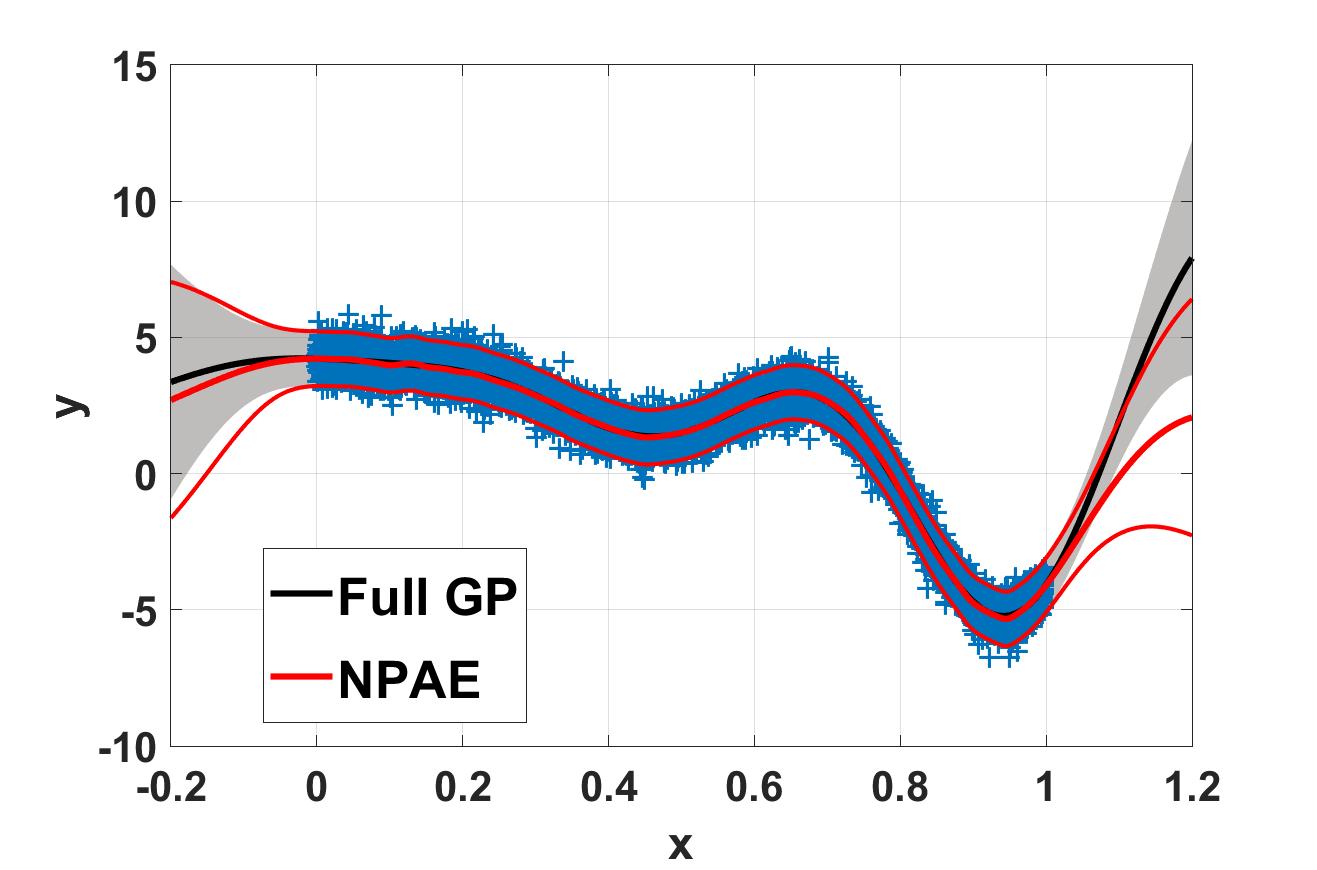}}\hfill
\subcaptionbox{GPOE}{\includegraphics[width=0.33\columnwidth]{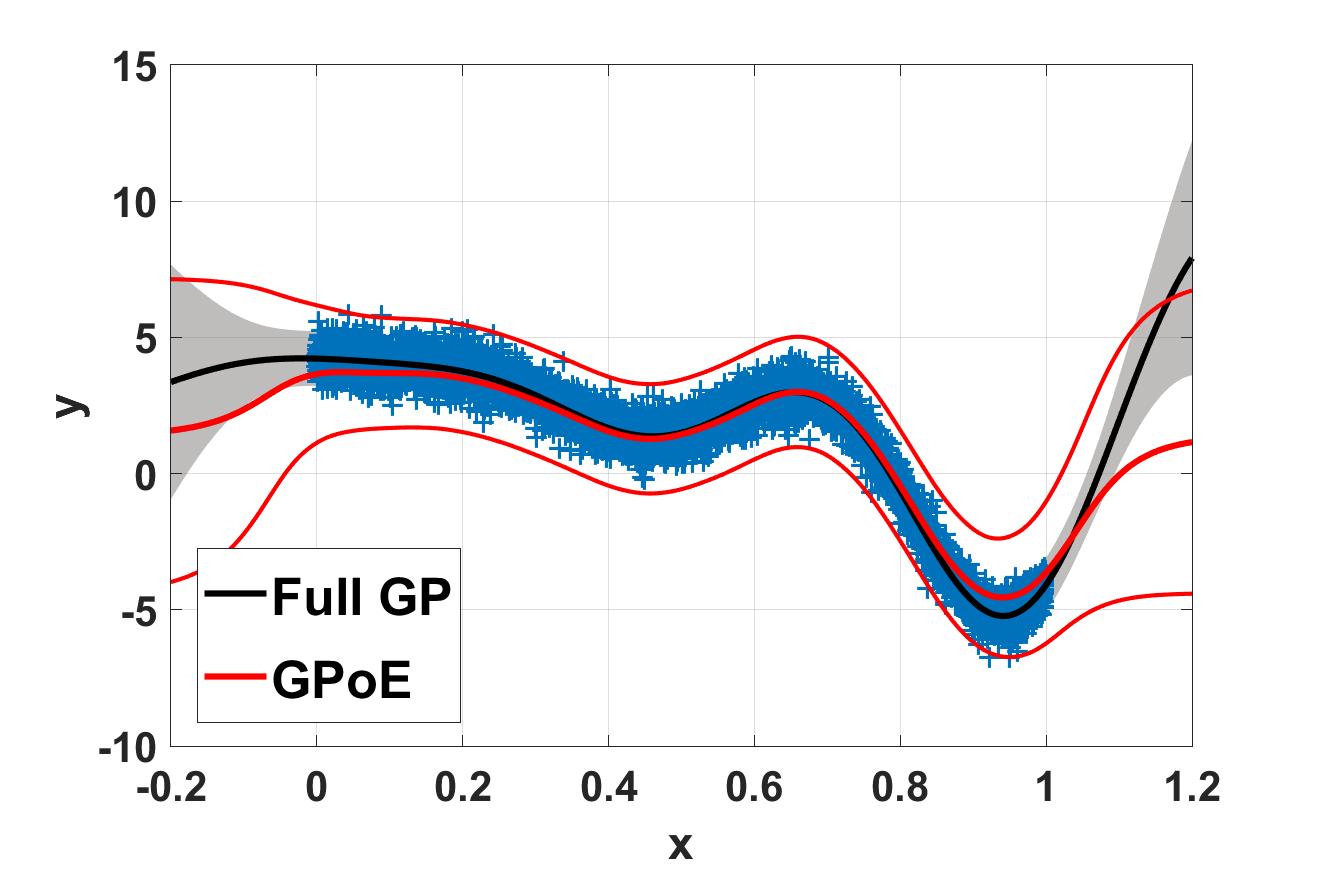}}\hfill
\subcaptionbox{RBCM}{\includegraphics[width=0.33\columnwidth]{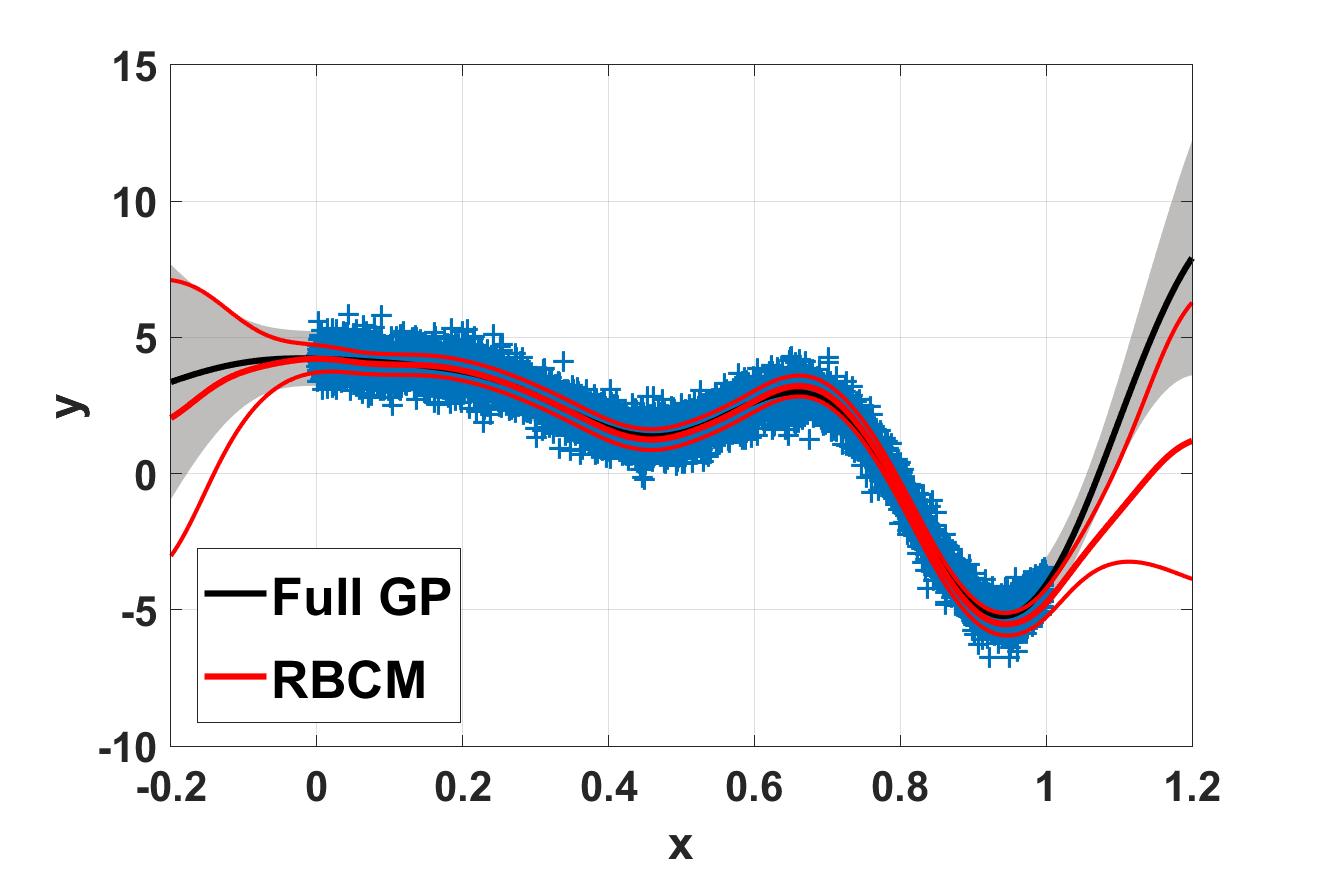}}\hfill
\subcaptionbox{GRBCM}{\includegraphics[width=0.33\columnwidth]{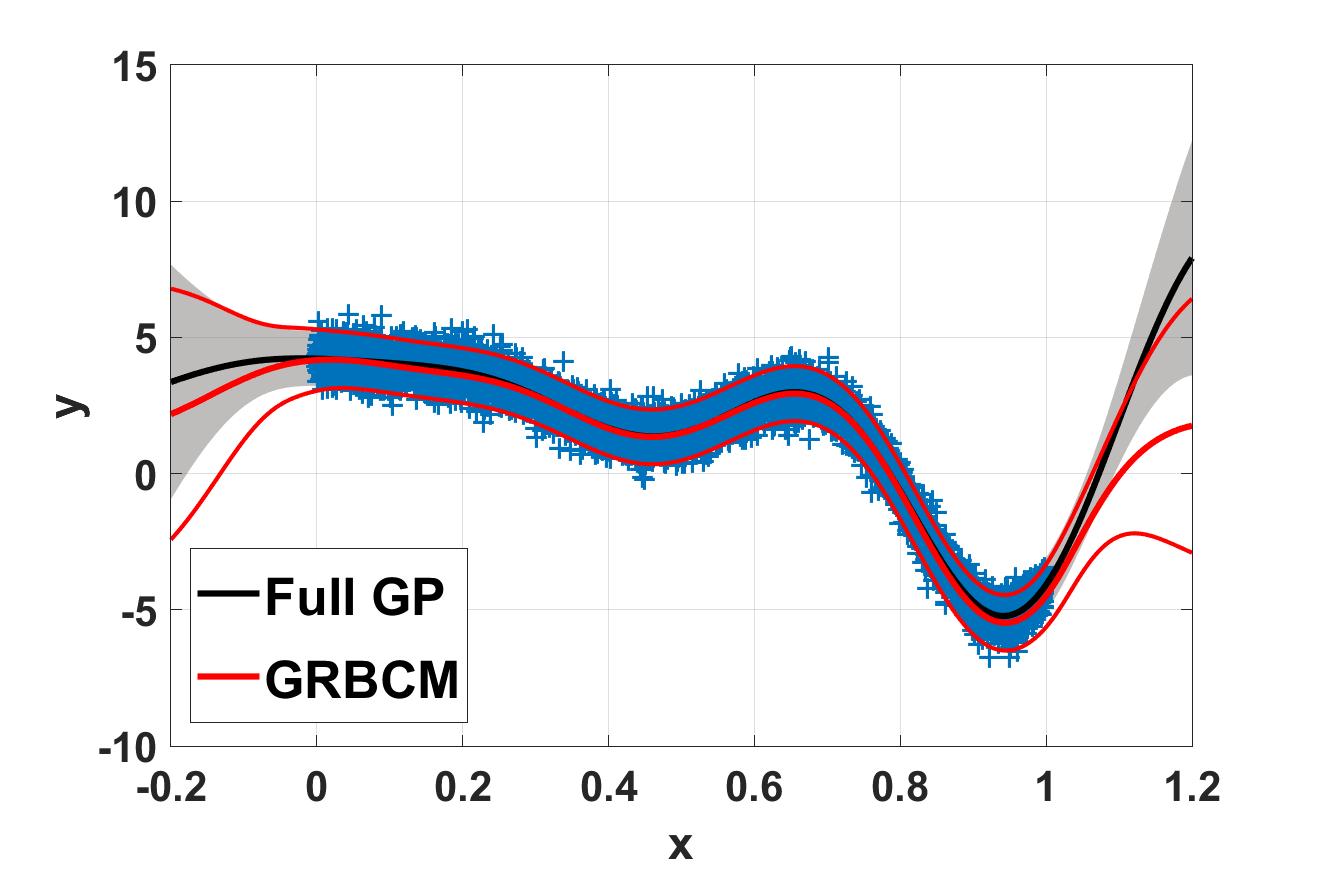}}\hfill
\caption{\textbf{99\% confidence interval} of different aggregation methods and full GP for $5 \times 10^3$ training points and partition size $m_0=250$ with $K$-means partitioning. The NPAE*(0.5) and NPAE*(0.8) results are based on $\alpha=0.5$ and $\alpha=0.8$, where $\lambda=0.1$.}
\label{fig.confidence_interval_fx}
\vskip -0.1in
\end{figure}

 Figure \ref{fig.confidence_interval_fx} depicts confidence intervals of different aggregation methods relative to the full GP from Section \ref{sec.4.1} with $M=20$. The NPAE, NPAE*(0.8), and NPAE*(0.5) produce reliable predictions when leaving the training data, i.e. for $ x\in[-0.2,0]$ and $x\in[1,1.2]$. Their estimated mean values and confidence intervals are highly accurate, while the other methods show significant deviation from the full GP, e.g. see the mean vector of CI-based method in $x \in [-0.2,0.2]$ and $x \in [0.8,1.2]$ of figure \ref{fig.confidence_interval_fx}. Both NPAE*(0.5) and NPAE*(0.8) return impressive results which means their NPAE approximations are acceptable. The main drawback of GRBCM is its mean value, especially for $x \in [0.8,1.2]$. However, its confidence intervals are much better than GPOE and RBCM methods. The results also fit the previous expectations about the GPOE and RBCM, RBCM provide overconfident results while GPOE is conservative.

\begin{figure}[htb!] 
\centering
\subcaptionbox{MAE\label{fig:appendix_mae}}{\includegraphics[width=0.33\columnwidth]{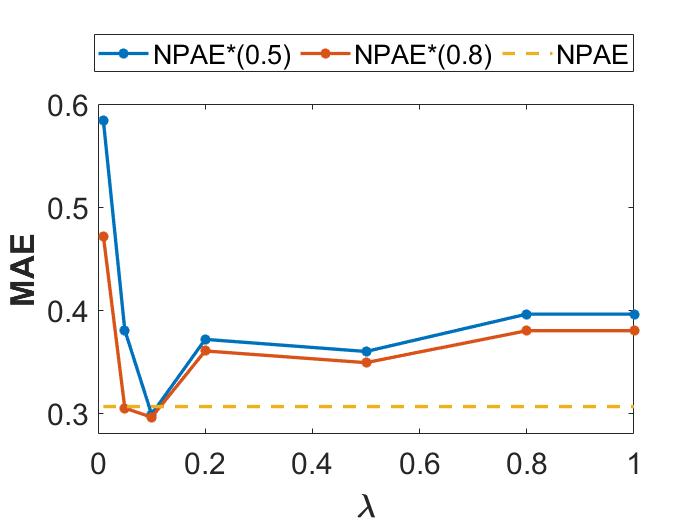}}\hfill
\subcaptionbox{SMSE\label{fig:appendix_smse}}{\includegraphics[width=0.33\columnwidth]{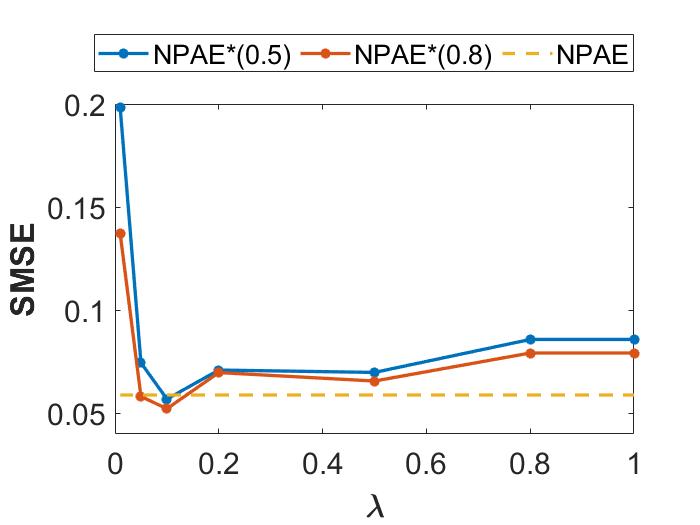}}\hfill
\subcaptionbox{MSLL\label{fig:appendix_msll}}{\includegraphics[width=0.33\columnwidth]{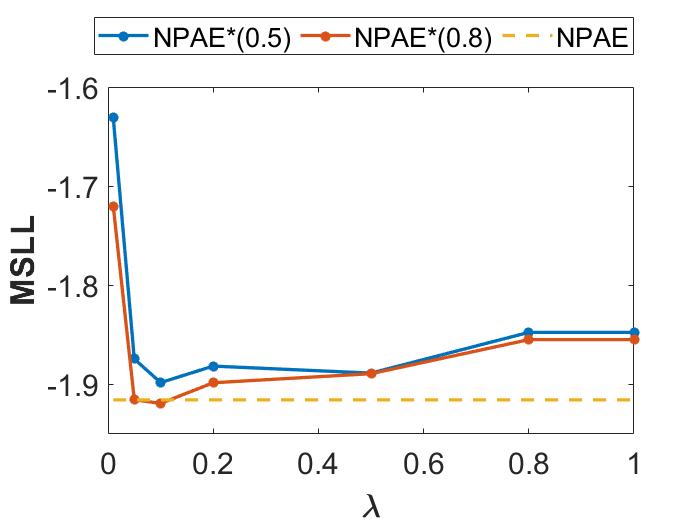}}\hfill
\caption{ MAE, SMSE and MSLL of NPAE* method with $\alpha=0.5$ and $\alpha=0.8$ for different values of the penalty term $\lambda$ .}
\label{fig.lambda}
\end{figure}

Finally, Figure \ref{fig.lambda} presents the varying of the sparsity parameter $\lambda$ to see how it affects the performance of NPAE*. It depicts the prediction quality measures for different $\lambda$ values for $5 \times 10^3$ observations of the synthetic data set in \eqref{f_x} and $M=20$. In this experiments, NPAE*(0.5) and NPAE*(0.8) are used, which are related to $\alpha=0.5$ and $\alpha=0.8$, respectively. As the figures \ref{fig:appendix_mae}, \ref{fig:appendix_smse} and \ref{fig:appendix_msll} show, small and large values (i.e. smaller than 0.05 or larger than 0.5) lead to slightly poor results but for values between 0.05 and 0.5, the network shows stable results. For the small $\lambda$ value, the graph is dense with more edges while large $\lambda$ leads to a very sparse network with only nodes and very few edges. Therefore, they do not provide appropriate results. The figures demonstrate $\lambda \approx 0.1$ is an appropriate choice to do the expert selection step.

A similar analysis can be found in Figure \ref{fig.importance} for $\alpha$ when $M=20$. The blue lines show the prediction quality of NPAE* when the value of the $\alpha$ changes ( $\alpha=1$ shows the conventional NPAE method). We can see that when for $\alpha \geq 0.5$, the blue lines are almost horizontal, which means for values in this range, NPAE* returns a very close prediction to the standard NPAE. By increasing the $\alpha$, the computational cost of NPAE* raises and approaches the NPAE's cost.

\section{Experimental Results: \textit{Pumadyn} Data set}\label{appendix:Pumadyn}

In this section we present the full experimental results for \textit{Pumadyn} data set in Section \ref{sec:4.2.1}. We consider the GPoE, RBCM, GRBCM, and NPAE with random and K-means partitioning. For the GPoE, RBCM and NPAE, we evaluate the proposed expert selection strategy with $\alpha= 0.8$ for GPoE and RBCM and $\alpha = 0.5$ and $\alpha = 0.8$ for NPAE. We used both disjoint and random partitioning to divide the data set into 7, 15, and 20 subsets.

\begin{table*}[hbt!]
\caption{SMSE and MSLL of different baselines in \textit{Pumadyn} data set for \textbf{K-means} partitioning strategy.}
\label{table.Pumadyn_kmeans_2}
\vskip -2 mm
\footnotesize
\begin{center}
\begin{small}
\begin{sc}
\resizebox{\textwidth}{!}{%
\begin{tabular}{lccccccr}
\toprule
\multirow{1}{*}{Number of Experts} &
\multicolumn{2}{c}{M=7} &
\multicolumn{2}{c}{M=15} & 
\multicolumn{2}{c}{M=20} & \\
\toprule
Model &  SMSE  & MSLL& SMSE  & MSLL & SMSE  & MSLL  \\
\midrule
GPoE   & 0.0475 & -1.5213 &  0.0483& -1.5166 & 0.0499& -1.490 &  \\
GPoE*    & 0.0473 & -1.5299 & 0.0477& -1.5213 & 0.0497&-1.493 & \\
RBCM     & 0.0474 & 2.2199 & 0.0478 & 1.1224&  0.0495 & 4.4283 &  \\
RBCM*     & 0.0473 & 1.5514 &  0.0474 & 0.3949 &  0.0493 & 3.0198 & \\
GRBCM      & 0.0476 & -1.5164 & 0.0499 &-1.4949 &  0.0494&-1.5029 & \\
\midrule
NPAE*(0.5)    &  0.0468&-1.5326  &  0.0470 & -1.5285 & 0.0483 &-1.5179 &  \\
NPAE*(0.8)    & 0.0465 &-1.5362 & 0.0468 &  -1.531 & 0.0480 & -1.5202  & \\
NPAE  &  0.0464 & -1.5367 & 0.0466& -1.536 &  0.0476 & -1.5245 & \\
\bottomrule
\end{tabular}}
\end{sc}
\end{small}
\end{center}
\vskip -0.1in
\end{table*}

\begin{table*}[hbt!]
\caption{SMSE and MSLL of different baselines in \textit{Pumadyn} data set for \textbf{Random} partitioning strategy.}
\label{table.Pumadyn_kmeans}
\vskip -2 mm
\footnotesize
\begin{center}
\begin{small}
\begin{sc}
\resizebox{\textwidth}{!}{%
\begin{tabular}{lccccccr}
\toprule
\multirow{1}{*}{Number of Experts} &
\multicolumn{2}{c}{M=7} &
\multicolumn{2}{c}{M=15} & 
\multicolumn{2}{c}{M=20} & \\
\toprule
Model &  SMSE  & MSLL& SMSE  & MSLL & SMSE  & MSLL  \\
\midrule
GPoE   & 0.0485 & -1.511 & 0.0488 & -1.5126 & 0.05 & -1.4789 \\
GPoE*    &  0.048& -1.52 & 0.0485&-1.518 & 0.0497 & -1.4802\\
RBCM     & 0.0534 & -0.0739 & 0.0485 & 2.9205 & 0.0495 & 3.4551 \\
RBCM*     & 0.052 & -0.3986 & 0.0482 &  1.7881 &  0.0492 &  2.2147\\
GRBCM      &  0.0498 & -1.5056  &   0.0507 & -1.502 &  0.0513 & -1.4904 \\
\midrule
NPAE*(0.5)    &  0.0478 & -1.525  &   0.0477 & -1.5265 &  0.0488 & -1.5039  \\
NPAE*(0.8)    &  0.0469 & -1.535  &  0.0474 & -1.530  &   0.0480 & -1.5103 \\
NPAE  &  0.0473 & -1.534 & 0.0472 & -1.5344 &    0.0478 & -1.5177 \\
\bottomrule
\end{tabular}}
\end{sc}
\end{small}
\end{center}
\vskip -0.1in
\end{table*}

\section{GRBCM Model with Expert Selection }\label{sec:GRBCM*}

The expert selection strategy can be easily extended to the GRBCM model. The modification in the GRBCM model, say GRBCM*, consists of the following two tasks:
\begin{enumerate}[(i)]
  \item excluding the weak experts based on the procedure explained in Section \ref{sec:3.1}
  \item choosing $\mathcal{I}_{i_1}$ (i.e., the top-most expert) in the sorted importance set $\mathcal{I}$ as the global communication expert (see Definition \ref{def.3}).
\end{enumerate}

The global communication expert $\mathcal{D}_b$ in GRBCM* can be selected via sorted importance set $\mathcal{I}$ in Definition \ref{def.3}, because it contains expert interactions and the intensity of those interactions.
The GRBCM* maintains the desired asymptotic properties of the original GRBCM, i.e. it leads to consistent estimation of the full Gaussian process at the presence of the assumption (i) and (ii) in Proposition \ref{prop:predictive_distribution} when $n \to \infty$. These modifications lead to a better predictive distribution at the end. While the first step eliminates the negative effects of weak experts in the final aggregation, the second step chooses the most reliable (i.e., interconnected) expert in the related GGM  as a global communicator. Note that the conventional GRBCM approach does not propose a strategy for selecting the global communication expert. Since this expert has a crucial impact on the final aggregation, the most interconnected expert is an appropriate choice for the global communication. 

\section{Experimental Results: Large Scale Data sets}\label{appendix:large_scale}
\begin{figure}[htb!] 
\centering
\subcaptionbox{\label{fig:appendix_protein}}{\includegraphics[width=0.33\columnwidth]{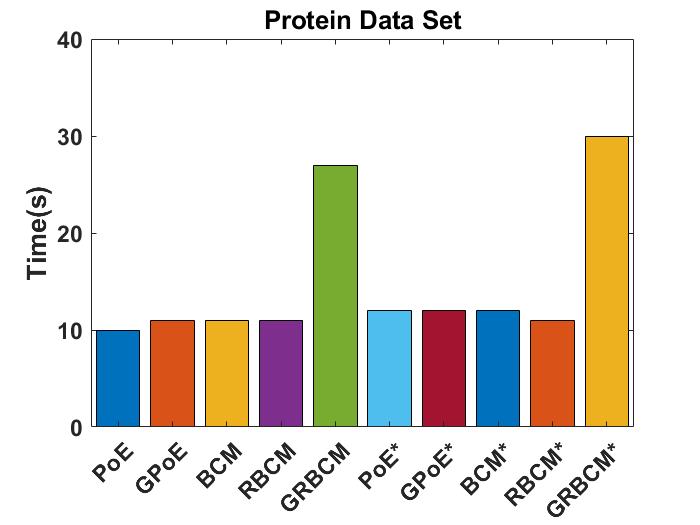}}\hfill
\subcaptionbox{\label{fig:appendix_sacros}}{\includegraphics[width=0.33\columnwidth]{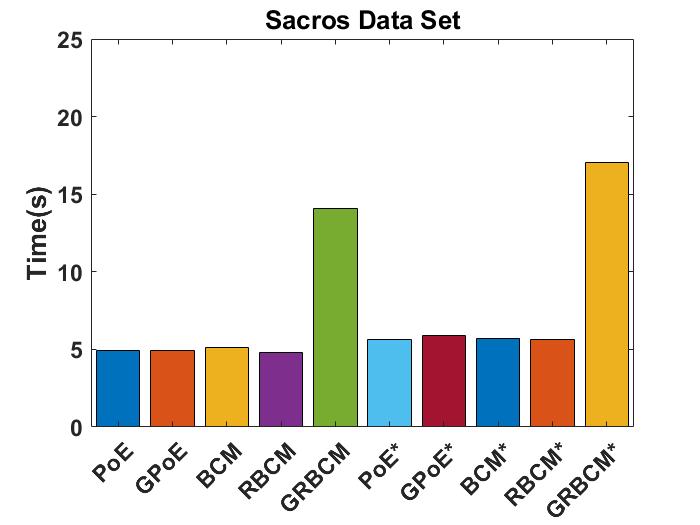}}\hfill
\subcaptionbox{\label{fig:appendix_song}}{\includegraphics[width=0.33\columnwidth]{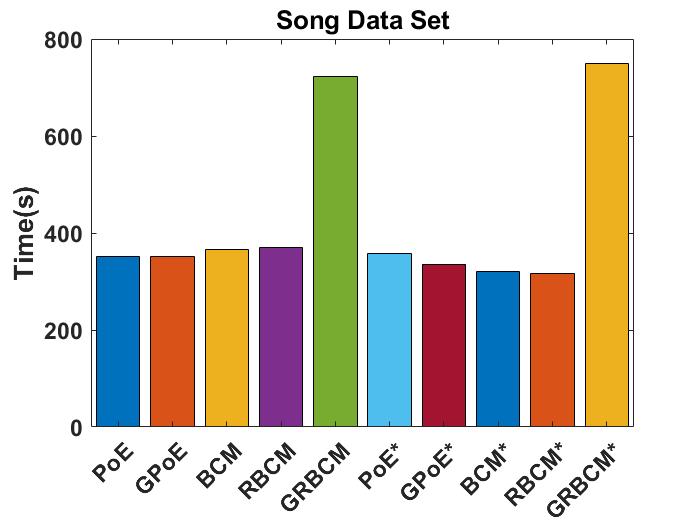}}\hfill
\caption{ Running Time of CI-based aggregation methods on \textit{Protein}, \textit{Sacros}, and \textit{Song} data sets.}
\label{fig.appendix_time_CI}
\end{figure}

Table \ref{table.CI} depicts the prediction quality of the CI-based aggregation method, i.e. (G)PoE, (R)BCM, and GRBCM. The consistent methods have been discussed in Table \ref{table.kmeans}. For a constant penalty term, $\lambda = 0.1$, the modified version of CI-based methods have been defined using $\alpha = 80 \%$ of the most important experts. The (G)PoE*, (R)BCM*, and GRBCM* are the expert selection version of the original (G)PoE, (R)BCM, and GRBCM methods. Table \ref{table.CI} shows a significant improvement after excluding the weak experts. Since the CI-based methods do not have appropriate asymptotic properties, they are not capable to outperform the NPAE which considers the experts' dependency. Figure \ref{fig.appendix_time_CI} presents the running time of the baselines on three large scale data sets in Table \ref{table.CI}. We can observe that the running time of both original and modified versions are of the same rate. The cost of the modified versions contain the cost of the GLasso (one time) and then the cost of aggregation on a smaller expert' set.

\begin{table*}[hbt!]
\caption{\textbf{Prediction quality} for various CI based baselines and their modified versions on \textit{Protein}, \textit{Sacros}, and \textit{Song} data sets.The table depicts SMSE and MSLL values for (G)PoE, (R)BCM, GRBCM, and their modified version after excluding the weak experts, i.e. (G)PoE*, (R)BCM*, and GRBCM* respectively. }
\label{table.CI}
\vskip -5 mm
\footnotesize
\begin{center}
\begin{small}
\begin{sc}
\resizebox{\textwidth}{!}{%
\begin{tabular}{lccccccr}
\toprule
\multirow{2}{*}{} &
\multicolumn{2}{c}{Protein} &
\multicolumn{2}{c}{Sacros} & 
\multicolumn{2}{c}{Song} & \\
\toprule
Model   & SMSE  & MSLL& SMSE  & MSLL & SMSE  & MSLL  \\
\midrule
PoE    & 0.8553 & 33.2404 & 0.045 & 2.724 &  0.952 & 70.325 \\
GPoE    & 0.8553& -0.082 & 0.045 & -1.183 &  0.952& -0.029 \\
BCM     &  0.3315 & -0.3329 &  0.007 & -2.359 & 2.660 & 3.438 \\
RBCM    & 0.3457 & -0.584 & 0.0045 & -2.413 & 0.847 & -0.006 \\
GRBCM   & 0.3477 & -0.613  & 0.0037 & -2.642 & 0.836 & -0.0926  \\
\midrule
PoE*   & 0.8146 & 26.0748 & 0.038 & 1.139 & 0.943 & 55.476\\
GPoE*  & 0.8146 & -0.1146 & 0.0038 & -1.282 & 0.943 & -0.034\\
BCM*   & 0.324 & -0.3776 & 0.007 & -2.500 & 2.161 & 2.349 \\
RBCM*  & 0.3411 & -0.5953  & 0.005 & -2.520 & 0.842 & -0.022 \\ 
GRBCM* & 0.3452 & -0.623 & 0.0035 & -2.731 & 0.819 & -0.102\\
\bottomrule
\end{tabular}}
\end{sc}
\end{small}
\end{center}
\vskip -0.1in
\end{table*}

\end{document}